\documentclass[10pt,twocolumn,letterpaper]{article}

\usepackage{iccv}
\usepackage{times}
\usepackage{epsfig}
\usepackage{graphicx}
\usepackage{amsmath}
\usepackage{amssymb}
\usepackage{listings}
\usepackage[normalem]{ulem}
\usepackage[accsupp]{axessibility}

\usepackage[breaklinks=true,bookmarks=false,colorlinks,pagebackref=true]{hyperref}

\iccvfinalcopy

\ificcvfinal\fi

\usepackage[square,numbers, sort]{natbib}
\usepackage{caption}
\usepackage[noabbrev, capitalize]{cleveref}
\usepackage{graphicx}
\usepackage{multirow}
\usepackage{subcaption}
\usepackage{algorithm}
\usepackage{algorithmic}
\usepackage{float}
\usepackage{booktabs}
\usepackage{bbm}
\usepackage[normalem]{ulem}
\usepackage{amsthm}
\usepackage{lipsum}

\usepackage[dvipsnames]{xcolor}
\usepackage{tabularx}
\usepackage{makecell}
\usepackage{colortbl}

\Crefname{ALC@unique}{Line}{Lines}
\Crefname{definition}{Definition}{Def}

\theoremstyle{definition}
\newtheorem{definition}{Definition}[section]

\newcommand{\srcdata}{\mathcal{D}_s}
\newcommand{\tgtdata}{\mathcal{D}_t}
\newcommand{\dset}{\mathcal{D}}
\newcommand{\loss}{\mathcal{L}}
\newcommand{\tgtacc}{Acc_t}
\newcommand{\srcacc}{Acc_s}
\newcommand{\placc}{Acc_{pl}}

\newcommand{\expect}[2]{\mathop{\mathbb{E}}_{#1}\left[#2\right]}
\newcommand{\kldiv}[2]{D_{KL}\left(#1||#2\right)}
\newcommand{\dtlarrow}{\xrightarrow[]{DTL}}

\newcommand{\ldtl}{\loss_{dtl}}
\newcommand{\llearn}{\loss_{learn}}

\newcommand{\gdtl}{\nabla\loss_{dtl}}
\newcommand{\glearn}{\nabla\loss_{retain}}

\begin{document}


\title{Disposable Transfer Learning for Selective Source Task Unlearning}
\author{Seunghee Koh$^1$ \quad Hyounguk Shon$^1$ \quad Janghyeon Lee$^2$\quad Hyeong Gwon Hong$^1$ \quad Junmo Kim$^1$\\
$^1$ Korea Advanced Institute of Science and Technology, South Korea \\
$^2$ LG AI Research, South Korea\\
{\tt\small \{seunghee1215, hyounguk.shon, honggudrnjs, junmo.kim\}@kaist.ac.kr}\\
{\tt\small janghyeon.lee@lgresearch.ai}
}

\maketitle
\ificcvfinal\thispagestyle{empty}\fi

\begin{abstract}
Transfer learning is widely used for training deep neural networks (DNN) for building a powerful representation. Even after the pre-trained model is adapted for the target task, the representation performance of the feature extractor is retained to some extent. As the performance of the pre-trained model can be considered the private property of the owner, it is natural to seek the exclusive right of the generalized performance of the pre-trained weight. To address this issue, we suggest a new paradigm of transfer learning called disposable transfer learning (DTL), which disposes of only the source task without degrading the performance of the target task. To achieve knowledge disposal, we propose a novel loss named Gradient Collision loss (GC loss). GC loss selectively unlearns the source knowledge by leading the gradient vectors of mini-batches in different directions. Whether the model successfully unlearns the source task is measured by piggyback learning accuracy (PL accuracy). PL accuracy estimates the vulnerability of knowledge leakage by retraining the scrubbed model on a subset of source data or new downstream data. We demonstrate that GC loss is an effective approach to the DTL problem by showing that the model trained with GC loss retains the performance on the target task with a significantly reduced PL accuracy.
\end{abstract}

\begin{figure}[t]
     \centering
     \includegraphics[width=\linewidth]{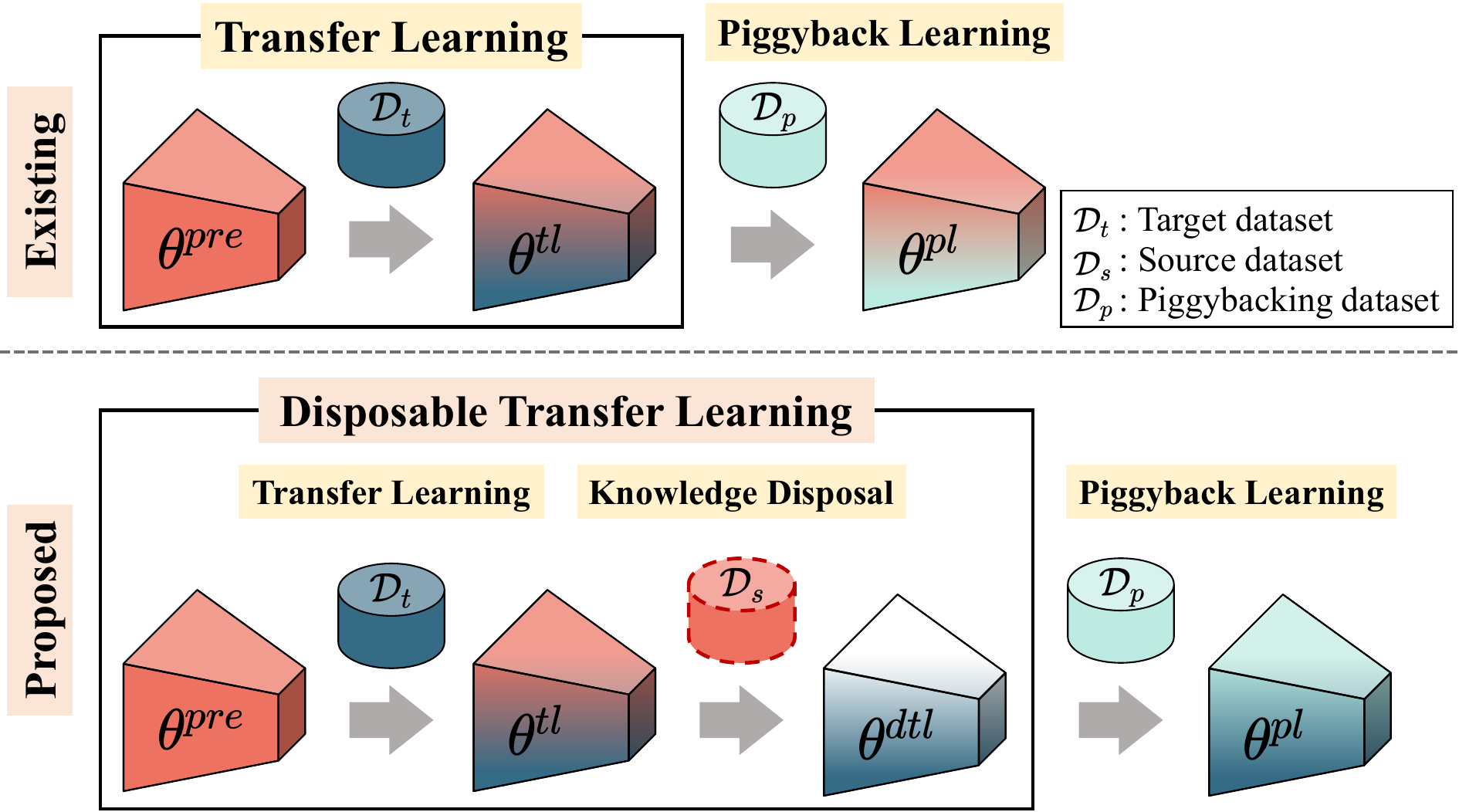}
     \caption{Illustration of the proposed Disposable Transfer Learning (DTL) framework. DTL extends the existing Transfer Learning (TL) paradigm with an additional knowledge disposal stage that scrubs off the prior knowledge irrelevant to the target task. The goal of DTL is to prevent Piggyback Learning (PL) which maliciously exploits the representation performance of a pre-trained model for a piggyback task by simply performing an extra fine-tuning step on top of the published transfer-learned model.}
     \label{fig:dtl-phases}
\end{figure}

\section{Introduction}

Transfer learning{~\cite{transfer_learning}} (TL) is one of the bedrocks in the success of deep neural networks (DNNs). The core idea of TL is to build a strong generic model that can adapt to a broad range of downstream tasks with much less amount of data. The scale of data collection for pre-training becomes the key objective to build a model with competitive performance on their target tasks~{\cite{BiT}}. Nowadays, a lot of organizations are interested in collecting an internal dataset to build their own generic model.

For example, JFT-300M~{\cite{jft1, jft2, jft3}}, a large-scale private dataset exclusively available to Google, is used for pre-training to reach the state-of-the-art performance of downstream target tasks~{\cite{KD_JFT, jft_3b}} with relatively small target datasets. IG-3.5B-17k~{\cite{fair_dataset}}, an internal Facebook AI Research dataset, is also used to train a weakly supervised model. Such datasets and trained models have a high economic value in that collecting data and training a model is time-consuming and costly, and a proficient model is readily adaptable to various commercial services.

\begin{figure}[t]
     \centering
     \includegraphics[width=0.8\linewidth]{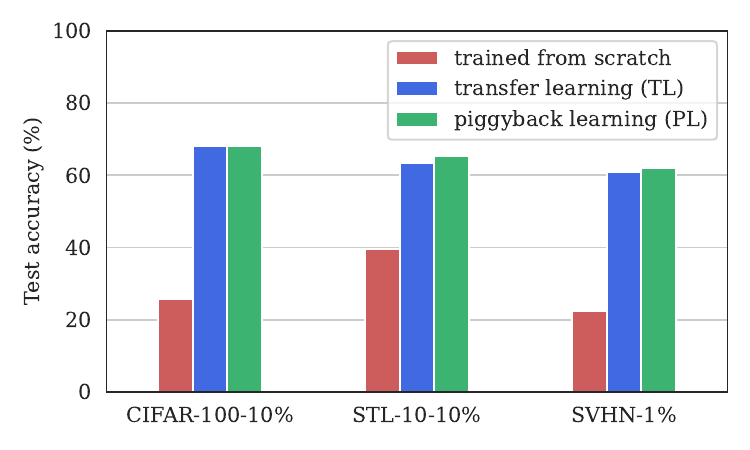}
    \caption{Performance comparison between models trained from scratch, models trained by transfer learning (TL), and models trained by piggyback learning (PL). The horizontal axis indicates the datasets used for performance evaluations. PL model (green) achieves performance comparable to TL model (blue) and both models perform much better than the models trained from scratch (red). For both TL and PL, CIFAR-100 is used as the source task. For PL, CIFAR-10-1\% is additionally used as the target task before the model is piggybacked.}
    \label{fig:finetuning_over_finetuning}
\end{figure}

However, those private properties are exposed to unauthorized customizing when released. Specifically, we denote piggyback learning (PL) (\cref{fig:dtl-phases}) as a kind of extra fine-tuning on other downstream tasks for leveraging the benefits of the transfer-learned model with much less effort. As shown in \cref{fig:finetuning_over_finetuning}, the performance of TL (blue) and PL (green) on the downstream tasks is comparable, and is considerably improved over the model trained from scratch (red). In other words, it enables anyone to exploit the pretext knowledge even when one does not have access to the pre-trained model by accessing the released transfer-learned model. PL is profitable to those free riders, so they may launch a new service or product by just exploiting the proficient model. It conflicts with the model owner's interest.

To alleviate this potential risk, we propose a novel TL paradigm that temporarily utilizes and then \emph{disposes of} the source task knowledge after transfer learning, coined \emph{Disposable Transfer Learning} (DTL). DTL aims to protect the exclusive license of generic performance on the internal pre-training data while achieving a powerful downstream performance.

To address the DTL problem, we propose a novel loss function that scrubs the source task knowledge, named \emph{Gradient Collision loss} (GC loss). GC loss guides the model towards abnormal convergence on the source task by minimizing inner-products between sample gradients. GC loss deals with a non-typical unlearning problem where the scale of data to be unlearned is much larger and the dataset to be unlearned and the dataset to be retained are heterogeneous, whereas existing unlearning literature {\cite{machine_unlearning, Mixed_privacy, Eternal_sunshine, Linear_Filtration}} mainly focus on forgetting only a small portion of a single kind of training data.

After DTL, we measure the model's susceptibility to unwanted PL using \emph{Piggyback Learning accuracy} (PL accuracy). We define the PL accuracy of a model as the test accuracy measured by learning an additional piggyback task. A low PL accuracy indicates that the model successfully unlearned the source knowledge so that it is resistant to a small portion of source re-training or fine-tuning on other downstream tasks. We will show the importance of PL accuracy as a measurement of unlearning for validating knowledge disposal.

We demonstrate that the model scrubbed with GC loss retains the target performance while effectively preventing the exploitation of the performance of the pre-trained model. To the best of our knowledge, DTL with GC loss is the first work in making transfer learning and unlearning compatible.

Our main contributions are summarized as follows:
\begin{itemize}
\item We propose a novel forgetting problem, DTL, in which we try to dispose of the generalization power of a pre-trained model while adapting the model only to a specific target downstream task.
\item We propose GC loss, which is a novel loss that achieves knowledge disposal of the source task. We also provide an extremely efficient implementation of GC loss that also allows for distributed training.
\item We propose PL accuracy as an evaluation metric that can estimate the performance of a DTL model.
\end{itemize}

\section{Related Work}
\subsection{Unlearning in deep learning}
Unlearning is a training mechanism to controllably forget specific data from the knowledge of a DNN and is gaining attention in the deep learning community owing to increased public awareness of digital privacy. The existing methods come with settings to bypass restrictions of the non-convex and stochastic nature of DNN, which make the perfect unlearning almost impossible.~{\cite{machine_unlearning}} combined distributed training and ensemble training to reduce retraining time excluding samples to be unlearned. Class-wise unlearning had been attempted~{\cite{Linear_Filtration}}, but it only removed a class from the classifier output, not from the model parameters. {\cite{Eternal_sunshine}} tried to scrub a subset of a class or an entire class with the assumption that the size of the dataset to be forgotten is much smaller than the dataset to be retained, and it uses the mean squared error loss to make the model partially convex to guarantee the perfect forgetting from the model. {\cite{Forgetting_outside}} approximated the weight difference using Neural Tangent Kernel theory~{\cite{NTK}}. {\cite{Mixed_privacy}} approximated the model to a linear model, with pre-determining a subset of the training dataset not to be unlearned and using it in pre-training.

\subsection{Gradient-based learning methods} 
The gradient vector of the learned model on specific input data characterizes the relationship between the model and input. In representation learning, a gradient of the learned model can be used as a feature for downstream tasks {\cite{Gradients_as_features}}. In continual learning, \cite{GEM, A-GEM} take advantage of gradient vectors to reduce catastrophic forgetting of a previously learned task. Also, it has been used to select samples for episodic memory{~\cite{Gradient_sample_selection}}. The algorithm for composing the episodic memory aims to minimize the cosine similarity of the gradient for each sample pair in candidate memory, and the authors have shown that the solution of the minimization problem is consequently maximizing the variance of the gradient of selected samples.

\subsection{Readout function}
\label{sec:readout}
Readout functions~{\citep{Mixed_privacy, Eternal_sunshine, Linear_Filtration}} are used to test whether a model has been successfully unlearned. Entropy, retraining time, or the success rate of membership inference attacks (MIAs) are usually used for the readout function. Entropy is used to measure the increase in uncertainty of the model after unlearning. This is based on the assumption that the prediction of a model gets less confident as it forgets the knowledge of interest. Retraining time quantifies the number of training steps required to restore the previous performance. MIA is a kind of attack method to detect whether a given sample was used in training a model, concerning the information leakage issue of training data from a trained model. Black-box MIAs only observe the input-output relationship, and white-box attacks fully exploit the model's architecture, weight, and gradient. We report the success rate of MIA against unlearned models using the strategies reported and implemented in \cite{AdvDistance}. Note that we use PL accuracy as the readout function for knowledge disposal.

\section{Method}
\textbf{Notations} Let $\dset$ be a dataset for a classification task with input space $\mathcal{X}$ and label space $\mathcal{Y}$.
We consider a DNN model $P(y|x;\theta)$ parameterized by $\theta$ that takes in an input $x$ and predicts a categorical probability distribution.
We denote a training procedure as $\theta^{out}=\textsc{Train}_\mathcal{A}\left(\theta^{in}, \mathcal{L}(\theta; \dset)\right)$ which represents a model learned from an initialization $\theta^{in}$, an objective function $\mathcal{L}(\theta; \dset)$, and a training scheme $\mathcal{A}$. For example, $\mathcal{A}$ can be a stochastic gradient descent optimizer with decaying learning rate scheduling.

\subsection{Disposable transfer learning}
\label{sec:dtl}
Disposable transfer learning (DTL) is a training paradigm for selective unlearning of the source task upon completion of transfer learning. As described in {\cref{fig:dtl-phases}}, it consists of two stages: the transfer learning (TL) stage and the knowledge disposal stage of the source data.

DTL is conducted by the owner of a private source data $\srcdata$, so $\srcdata$ is to be unlearned and accessible during unlearning. Also, the size of the target dataset $\tgtdata$ is much smaller, \ie, $|\srcdata| \gg |\tgtdata|$, such that transfer learning is required to achieve competitive target task performance.

\textbf{Transfer learning stage}\quad The transfer-learned model $\theta^{tl}$ is obtained by pre-training and fine-tuning. The model is initially pre-trained on a source dataset $\srcdata$ from scratch, and the output model $\theta^{pre}$ is then easily adaptable to the target task. 
Then $\theta^{pre}$ is fine-tuned to get 
$\theta^{tl}=\textsc{Train}_\mathcal{A}\left(\theta^{pre}, \mathcal{L}(\theta; \dset_t)\right)$. Due to the small scale of $\tgtdata$, fine-tuning has a reasonable training cost and the fine-tuned weight $\theta^{tl}$ is not perturbed significantly from $\theta^{pre}$. 

\textbf{Knowledge disposal stage}\quad Knowledge disposal stage is the last stage of DTL for building $\theta^{dtl}$ by disposing of the source task from the transfer-learned model $\theta^{tl}$ with the DTL loss $\mathcal{L}_{DTL}$, which we denote as $\theta^{dtl}=\textsc{Train}_\mathcal{A}\left(\theta^{tl}, \mathcal{L}_{DTL}(\theta; \dset)\right)$.

For a successful disposable \emph{transfer learning} where good generalization to the target task is one of the key factors, $\theta^{dtl}$ has to retain the performance on the target task of $\theta^{tl}$. Simultaneously, it is essential to dispose of the source performance not to be recovered through PL, as discussed in \cref{sec:pl}. We formulate those factors into the retaining loss and the unlearning loss by combining them into a single objective as \cref{eq:total_dtl_loss}. Here, $\lambda$ is a scalar hyperparameter that controls the level of unlearning.
\begin{equation}
    \label{eq:total_dtl_loss}
    \mathcal{L}_{DTL}(\theta) = (1-\lambda)\cdot\mathcal{L}_{retain}(\theta) + \lambda\cdot\mathcal{L}_{unlearn}(\theta)
\end{equation}
\begin{figure}[t]
  \centering
  \includegraphics[width=0.95\linewidth]{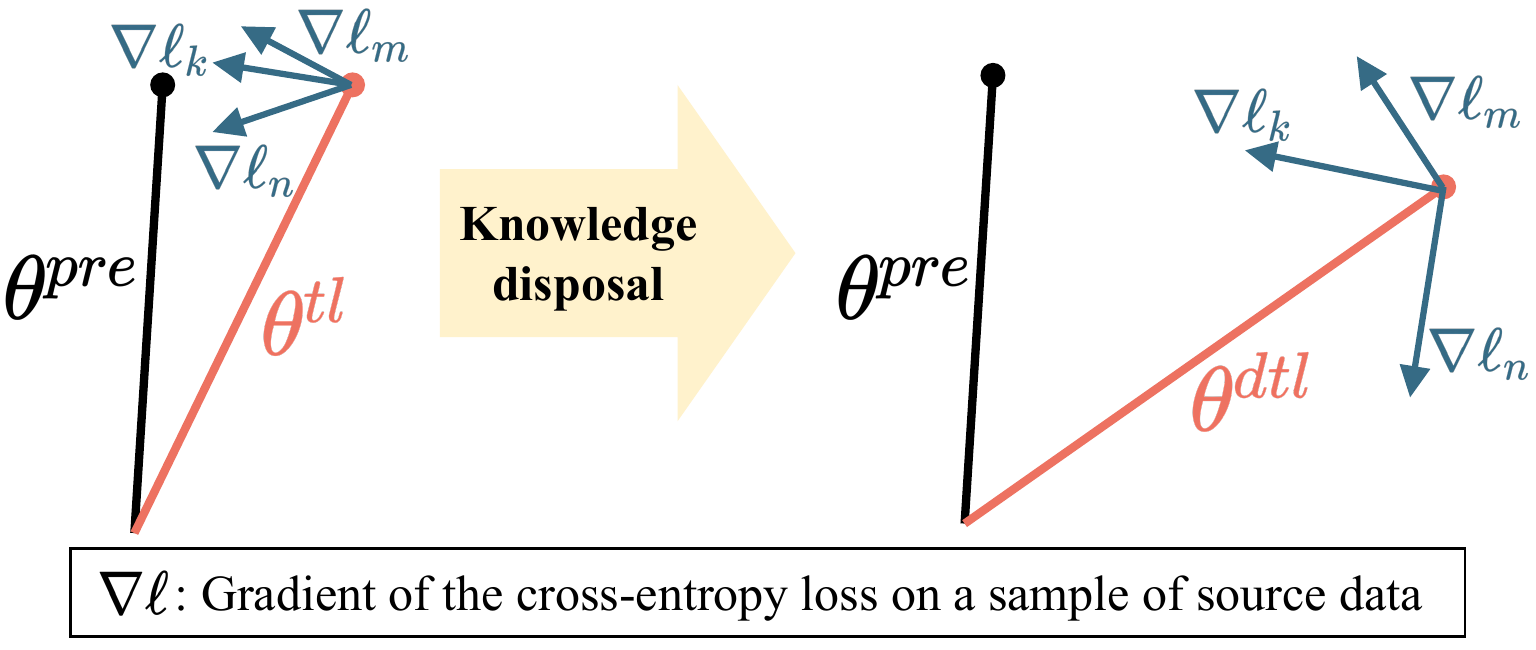}
  \caption{A conceptual diagram of gradients after transfer learning (left) and knowledge disposal (right). The gradients from the source data examples are marked as blue arrows. As depicted in ~\cref{fig:dtl-phases}, $\theta^{pre}$, $\theta^{tl}$ and $\theta^{dtl}$ correspond to the pre-trained model, transfer-learned model, and disposable transfer-learned model, respectively.}
  \label{fig:loss_landscape}
\end{figure}
\begin{figure*}[t]
  \centering
  \includegraphics[width=\linewidth]{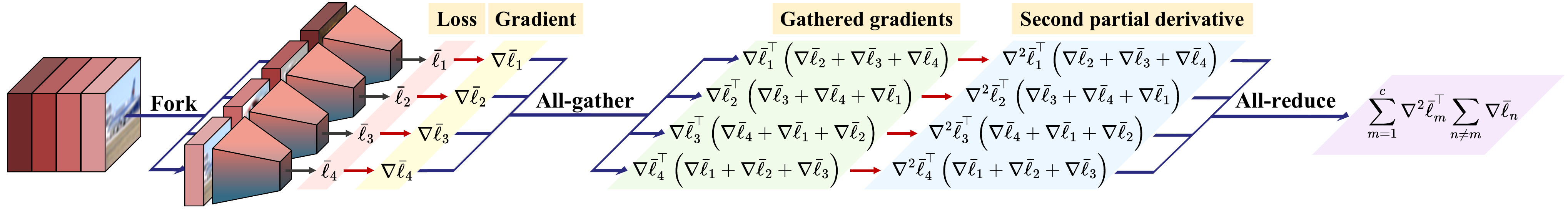}
  \caption{Illustration of distributed computation for GC loss. The computation of GC loss and its gradient can be optimized by rearranging \cref{eq:deriv_L_gc}. The rearrangement makes the HVP computation parallelized and enables distributed processing of the backward pass. The pseudo-code is reported in \cref{alg:ddp_loss_gcl}.}
  \label{fig:l_overfit_ddp}
\end{figure*}
\subsection{Retaining loss and unlearning loss}
\subsubsection{Retaining by knowledge distillation loss}
\label{sec:retain_loss}
To effectively retain the transferred knowledge on the downstream task, we adopt knowledge distillation loss (KD loss)~{\cite{KD_JFT}}. KD loss is for transferring knowledge between different models by setting the output of the teacher model to a soft target and minimizing the KL divergence of the soft target and the output of a student model, as formulated in \cref{eq:tgt_keeping_loss}. In this paper, we choose $\dset = \srcdata$ to prevent a risk of over-fitting due to the small size of $\tgtdata$.
\begin{equation}
    \label{eq:tgt_keeping_loss}
   \mathcal{L}_{retain}(\theta) = \expect{x \sim \dset}{ \kldiv{P(y|x;\theta^{tl})}{P(y|x;\theta)} }
\end{equation}
\subsubsection{Unlearning by gradient collision loss}
Consider a dataset of training examples $\dset=\{(x_i,y_i)\}_{i=1}^N$. The cross-entropy loss of a model $\theta$ on an example $(x_i,y_i)$ is denoted by $\ell(\theta; x_i,y_i)$, which is abbreviated to $\ell_i(\theta)$ when clear from the context. 

In the context of unlearning, a key concept is how the value of $\ell_n(\theta)$ changes when the model weights are updated to reduce $\ell_m(\theta)$. In gradient descent, the model weight is optimized to minimize the loss function by updating $\theta$ by $\Delta\theta$ which is $\nabla\ell(\theta)$ scaled by the learning rate $\eta$:
\begin{equation}
    \Delta \theta = -\eta\nabla\ell(\theta; x_m,y_m)
    \label{eq:delta_theta}
\end{equation}
The change of $\ell_n(\theta)$ over $\Delta \theta$ is approximated by its first-order Taylor expansion,
\begin{align}
    \Delta \ell_n(\theta) &= \ell_n(\theta + \Delta\theta) - \ell_n(\theta) \\
    &\simeq {\nabla\ell_n(\theta)}^\top\Delta\theta.
    \label{eq:delta_ell}
\end{align}
We get the following relationship between a pair of gradients on the loss by combining \cref{eq:delta_theta} and \cref{eq:delta_ell}.
\begin{equation}
    \Delta \ell_n(\theta) \simeq -\eta{\nabla\ell_m(\theta)}^\top \nabla\ell_n(\theta)
    \label{eq:taylor_loss}
\end{equation}
\cref{eq:taylor_loss} states that the value of $\ell_n(\theta)$ is affected by the relationship between $\nabla\ell_m$ and $\nabla\ell_n$. As illustrated in the left side of \cref{fig:loss_landscape}, the gradients of a pair of examples point in similar directions, $\ie$ the inner-product of the gradients is positive, the model update for reducing $\ell_m(\theta)$ also causes a decrease in $\ell_n(\theta)$. On the other hand, as in the right side of \cref{fig:loss_landscape}, the update for reducing the loss on a sample does not imply the other's learning or can even hinder the other's learning.
Hence, we claim that the behavior of the model to learn the knowledge can be hindered by colliding the gradient of the examples of the dataset, which corresponds to reducing the inner product of the gradient vectors. Finally, we propose the gradient collision loss (GC loss).
\begin{definition}[Gradient Collision loss]
The full gradient collision loss defined for a dataset $\dset$ of size $N$ is given by,
\begin{equation}
    \label{eq:gc_loss_ideal}
    \mathcal{L}_{gc}(\theta, \dset) = {\frac{1} {\binom{N} {2}}} \sum_{m \neq n}{\nabla\ell_m(\theta)^\top \nabla\ell_n(\theta)}.
\end{equation}
\end{definition}
 When the model is unlearned, each gradient vector changes direction as seen in the \cref{fig:loss_landscape} and they converge towards perpendicular angles which results in suppressed inner-product value. Note that GC loss does not have a zero minimum. The value is negative when the per-sample gradients point in opposing directions. In this case, a gradient descent step on one sample leads to a loss increase on another sample.

This ideal form aggregates every pair of possible combinations of $\ell_i$, however, the computation quickly becomes intractable as the size $N$ grows. Therefore, we calculate the inner products within a stochastically sampled mini-batch by dividing a mini-batch into $c$ number of chunks and colliding the gradients of the chunks. We calculate the unlearning loss for every chunk pair, which are $\binom{c}{2}$ number of pairs per mini-batch.

\begin{definition}[Stochastic Gradient Collision loss] We define a variant of GC loss for mini-batch training. The stochastic GC loss is the sum of inner products between pairs of chunk-averaged gradients. Here, $\nabla \bar{\ell}_i$ is the gradient averaged over the $i$-th chunk in a mini-batch, and $c$ is the number of chunks per mini-batch.

\begin{equation}
    \label{eq:grad_collision_batch}
    \mathcal{L}_{gc}(\theta, \dset) = {\frac{1} {\binom{c} {2}}} {\sum_{m \neq n} \nabla \bar{\ell}_m(\theta)^\top \nabla \bar{\ell}_n(\theta)}
\end{equation}
\end{definition}
We compute GC loss on $\srcdata$, for disposing of the source knowledge by guiding the gradient of source data colliding. We use \cref{eq:grad_collision_batch} as our unlearning objective along with a custom gradient computation for efficient training.

\noindent \textbf{Efficient training of GC loss}\quad
Training with a naive implementation of GC loss is expensive because it requires calculating the gradient of each chunk in series, summing up the inner product of all pairs of the gradient vectors, and then conducting back-propagation. 

For computational efficiency, we re-formulate the derivative of GC loss to a sum of Hessian-vector products (HVPs), as in \cref{eq:deriv_L_gc_parallel}.

\begin{align}
    \label{eq:deriv_L_gc}
    \nabla\mathcal{L}_{gc}
    &=\frac{1} {\binom{c} {2}}{\sum_{m \neq n}\nabla\left({\nabla\bar{\ell}_m}^\top \nabla\bar{\ell}_n\right)}\\
    &=\frac{1} {\binom{c} {2}}{\sum_{m=1}^c{{\nabla^2\bar{\ell}_m}^\top\sum_{m \neq n}\nabla\bar{\ell}_n}}
    \label{eq:deriv_L_gc_parallel}
\end{align}
This reduces repeated computation of intermediate vectors and parallelizes computing the derivative of each chunk and each HVP along the chunk axis for distributed processing across multiple GPUs as depicted in \cref{fig:l_overfit_ddp}.

A detailed description of the algorithm is provided in \cref{alg:ddp_loss_gcl}. We use \texttt{DistributedDataParallel} \citep{pytorch} primitives of PyTorch which provide communication for multiprocessing. 
First, each process calculates the chunk-wise gradient of cross-entropy loss (\cref{lst:line:grad}).
The gradients are aggregated across processes using \texttt{gather} operation to compute the GC loss (\cref{lst:line:detach_gather}). \texttt{detach} allows for calculating the partial derivative of gradient which is required for computing the HVP. Finally, the GC loss is then partially back-propagated, and we \texttt{reduce} the values to obtain the gradient of GC loss as \cref{eq:deriv_L_gc_parallel} to update the model parameters.

While a naive combinatorial implementation for GC loss has $\mathcal{O}(c^2)$ complexity, our re-formulation greatly reduces the cost to $\mathcal{O}(c)$. Note that this matches the complexity of typical loss functions. This is because our method only needs products between individual vectors against an averaged value, not pair-wise products, which enables data parallelism for multiprocessing. 
An extra cost is a backward-on-backward step for propagating the gradient of GC loss in ~\cref{lst:line:bwd_on_bwd}. 
Fortunately, this only requires a similar computational cost and memory footprint to a typical backpropagation.
\begin{algorithm}[t]
\caption{Pseudocode for GC loss, PyTorch-like}
\definecolor{codeblue}{rgb}{0.25,0.5,0.5}
\definecolor{codekw}{rgb}{0.85, 0.18, 0.50}
\lstset{
  backgroundcolor=\color{white},
  basicstyle=\fontsize{7.2pt}{7.2pt}\ttfamily\selectfont,
  columns=fullflexible,
  breaklines=true,
  captionpos=b,
  commentstyle=\fontsize{3.5pt}{3.5pt}\color{codeblue},
  keywordstyle=\fontsize{7.5pt}{7.5pt}\color{codekw},
  numbers=left,
  stepnumber=1,    
  firstnumber=1,
  numberfirstline=true,
  escapechar=|,
}

\begin{lstlisting}[language=python]
# Inputs: net, input, target, lr
ce_loss = cross_entropy(net(input), target)
ce_grad = grad(ce_loss, net.parameters()) |\label{lst:line:grad}|
chunk_grad = all_gather(ce_grad.detach()).sum(0) |\label{lst:line:detach_gather}|
gc_loss = dot_prod(ce_grad, chunk_grad)
gc_grad = grad(gc_loss, net.parameters()) |\label{lst:line:bwd_on_bwd}|
gc_grad = all_reduce(gc_grad)
optimizer_step(net.parameters(), gc_grad, lr)
\end{lstlisting}
\label{alg:ddp_loss_gcl}
\end{algorithm}

\subsubsection{Baseline unlearning methods}
\label{sec:unlearning_baseline}
As baselines, we consider three additional naive unlearning losses adopted from \cite{Eternal_sunshine}.
Random target fooling loss updates the model using cross-entropy loss with $\tilde{\dset}$, a dataset constructed from $\dset$ with random target labels.
This leads the model to memorize wrong answers so that the model unlearns the related knowledge.
\begin{equation}
    \label{eq:rand_loss}
    \mathcal{L}_{rand}(\theta; \dset) = \expect{(x,y)\sim \tilde{\dset}}{-\log P(y|x;\theta)}
\end{equation}

Uniform target fooling loss guides the model to output uniform distribution, therefore the model loses the ability to make any prediction. Here, $\mathcal{U}(y)$ is an uniform distribution over $\mathcal{Y}$.
\begin{equation}
    \label{eq:unif_loss}
    \mathcal{L}_{unif}(\theta; \dset) = \expect{x\sim \dset}{\kldiv{\mathcal{U}(y)}{P(y|x;\theta)}}
\end{equation}

Negative cross-entropy loss flips the learning signal by increasing the cross-entropy. The concept is that increasing loss through gradient ascent steps lets a model be forgotten.

\begin{equation}
    \label{eq:neg_loss}
    \mathcal{L}_{neg}(\theta; \dset) = \expect{(x,y)\sim \dset}{\log P(y|x;\theta)}
\end{equation}

We will compare the proposed unlearning loss $\mathcal{L}_{gc}$ against $\mathcal{L}_{rand}$, $\mathcal{L}_{unif}$, and $\mathcal{L}_{neg}$ on $\srcdata$ in \cref{sec:exp}.

\definecolor{Gray}{gray}{0.87}
\begin{table*}[t]
\centering
\resizebox{\linewidth}{!}{
\begin{tabular}{c | r r r r r | r r r r r| r r r r r}
\noalign{\smallskip}
\toprule
 \multirow{3}{*}{\large{Model} }& \multicolumn{5}{c|}{CIFAR-100 $\dtlarrow$ CIFAR-10-1\%}
  & \multicolumn{5}{c|}{CIFAR-100 $\dtlarrow$ STL-10-10\%}
 & \multicolumn{5}{c}{TinyImageNet$\dtlarrow$CIFAR-100-10\%}\\
\cmidrule{2-16}
& \multirowcell{2}
{\footnotesize{$\Delta\tgtacc$} \\ \footnotesize{vs TGT $\uparrow$}} & \multirowcell{2}{\footnotesize{$\Delta\tgtacc$} \\\footnotesize{vs TL $\uparrow$}}
& \multicolumn{3}{c|}{$\Delta\placc$ vs TL $\downarrow$}

& \multirowcell{2}
{\footnotesize{$\Delta\tgtacc$} \\ \footnotesize{vs TGT $\uparrow$}} & \multirowcell{2}{\footnotesize{$\Delta\tgtacc$} \\\footnotesize{vs TL $\uparrow$}}
& \multicolumn{3}{c|}{$\Delta\placc$ vs TL $\downarrow$}

& \multirowcell{2}
{\footnotesize{$\Delta\tgtacc$} \\ \footnotesize{vs TGT $\uparrow$}} & \multirowcell{2}{\footnotesize{$\Delta\tgtacc$} \\\footnotesize{vs TL $\uparrow$}}
& \multicolumn{3}{c}{$\Delta\placc$ vs TL $\downarrow$}

\\
\cmidrule{4-6}\cmidrule{9-11}\cmidrule{14-16}
&   & &\scriptsize{C100-10\%}& \footnotesize{S10-10\%} & \footnotesize{SV-1\%}
&   & &\scriptsize{C100-10\%}& \footnotesize{C10-1\%} & \footnotesize{SV-1\%}
&   & &\footnotesize{TIN-6\%}& \footnotesize{S10-10\%} & \footnotesize{C10-1\%}
\\
\midrule
RAND	&	+33.88	&	-1.93	&	-14.16	&	-3.00	&	+4.01 &	+22.89	&	-0.98	&	-11.27	&	-1.91	&	+6.42    & +28.63 & -2.20 & -9.98 & -1.90 & -4.68  \\
UNIF 	&	+36.29	&	+0.48	&	-13.04	&	-1.17	&	+7.66 &	+23.46	&	-0.41	&	-12.48	&	-2.55	&	+6.40   & +29.63 & -1.20 & -11.17 & -3.50 & -3.48 \\
NEG     &	+36.05	&	+0.24	&	-7.18	&	-0.09	&	+9.13    &	+21.28	&	-2.59	&	-10.40	&	-4.15	&	+9.65   & +28.80 & -2.03 & -11.98 & -2.77 & -3.87 \\
\rowcolor{Gray}
\footnotesize{GC (Ours)}	&	+33.84	&	-1.97	&	\textbf{-24.53}	&	\textbf{-7.47}	&	\textbf{-6.10}    &	+21.95	&	-1.92	&	\textbf{-22.99}	&	\textbf{-10.93}	&	\textbf{-4.19}	  & +29.19 & -1.64 & \bf -12.79 & \bf -4.88 & \bf -7.90 \\
\midrule
\emph{\footnotesize{Reference}} & 35.12    &   70.93   &	68.10	&	65.25	&	61.97    &   39.58   &   63.45	&	68.15	&	72.12	&	60.82   &
25.16 & 55.99 & 39.17 & 67.71 & 72.08 \\
\bottomrule
\noalign{\smallskip}
\end{tabular}
}
\caption{The transition of source, target, and piggyback learning accuracy for each DTL stage in CIFAR-100 $\dtlarrow$ CIFAR-10-1\%, CIFAR-100 $\dtlarrow$ STL-10-10\%, and TinyImageNet $\dtlarrow$ CIFAR-100-10\% experiments. The values indicate the difference between DTL and TL counterparts as percentage points. The proposed GC loss significantly outperforms unlearning baselines in achieving DTL. This provides control over selectively removing pretext knowledge from the model after learning the target task. \emph{Reference} indicates the base accuracy (\%) used to obtain the difference.
CIFAR-10/100, STL-10, SVHN, and TinyImageNet are abbreviated as C10/100, S10, SV, and TIN, respectively. 
} 
\label{tab:main_result}
\end{table*}

\subsection{Evaluation by piggyback learning accuracy}

\label{sec:pl}
In this section, we establish an evaluation protocol for DTL performance using piggyback learning accuracy (PL accuracy). While measuring the source task accuracy may seem like the most direct approach for evaluating knowledge disposal, we have found that it is less effective measure due to trivial solutions such as last-layer fooling. Specifically, if the source task accuracy is used as an unlearning metric, a trivial solution is to simply collapse the source classifier since there are separate classifiers for the unlearned task and the retained task in DTL.
Designing a proper evaluation protocol for DTL is crucial, as the ultimate goal is to prevent the model from learning an unknown piggyback task after transfer learning is completed. 
To address these challenges, we propose to benchmark DTL using PL accuracy.

\begin{definition}[Piggyback learning and piggyback learning accuracy]
\label{def:pl_acc}
We define Piggyback Learning (PL) as adapting a model $\theta^{0}$ on a piggyback task $\dset_p$ using a fine-tuning scheme $\mathcal{A}$. 
\begin{equation}
    \theta^{pl}=\textsc{Train}_\mathcal{A}\left(\theta^{0}, \mathcal{L}(\theta; \dset_p^{train})\right)
\end{equation}
where $\dset_p^{train}$ is the train split of $\dset_p$.
Piggyback learning accuracy $\placc(\theta^0)$ of a model $\theta^{0}$ is the test accuracy of $\theta^{pl}$ on the piggyback task. $\theta^0$ is set to either $\theta^{tl}$ or $\theta^{dtl}$ in our main experiments.
\end{definition}
We use multiple datasets for measuring PL accuracy. Unlike a typical test accuracy, which is measured on a certain dataset, our protocol estimates the performance by fine-tuning the model obtained from DTL and testing the performance on multiple datasets. 

The PL accuracy quantifies the susceptibility of a model to various kinds of possible downstream tasks. In other words, it measures the model's transferability as a pre-trained weight. 

When $\dset_p$ is substituted with the source dataset $\dset_s$, PL accuracy can be also used to estimate the recoverability of the scrubbed model.
Lower PL accuracy implies a lower risk of catastrophic leakage of generic performance because the difficulty of relearning depends on the amount of knowledge remaining in the unlearned model.

\section{Experiment}
\label{sec:exp}
\subsection{Experimental setting}

We denote a DTL training scheme as $\dset_s\dtlarrow\dset_t$ where $\dset_s$ is the source task and $\dset_t$ is the target task. We used CIFAR-10/100, STL-10, SVHN, and TinyImageNet \cite{dataset_cifar,STL10,svhn,Tinyimagenet} to construct the benchmarks. Additionally, we reduced the scale of the target training datasets by sub-sampling. The training sets are reduced from the original size by class-balanced random sampling. Sub-sampled datasets have their name suffixed by ``-$\gamma$\%'', where $\gamma$ is the sampling ratio. Reducing the dataset size sets heavier emphasis on transfer learning as TL brings more impact to the target task performance ($\tgtacc$). The unlearned models are evaluated by the target accuracy and PL accuracy ($\placc$). For measuring the PL accuracy, the datasets are subsampled. Columns marked with an up-arrow symbol ($\uparrow$) indicate that higher is better, while a down-arrow symbol ($\downarrow$) indicates the opposite. We used ResNet-18{~\citep{resnet}} architecture for all experiments.

\subsection{Baseline methods}
The most naive baseline is {\small TGT} model, which is trained only with $\tgtdata$ from scratch. In addition, we compared the effect of the baseline unlearning losses in \cref{sec:unlearning_baseline} by conducting the knowledge disposal stage from the transfer-learned model ({\small TL}). We named each model following the name of unlearning loss. The model unlearned using the proposed GC loss (\cref{eq:grad_collision_batch}) is marked with {\small GC}.
We use random target fooling loss ({\small RAND}, \cref{eq:rand_loss}), uniform target fooling loss ({\small UNIF}, \cref{eq:unif_loss}), and negative cross-entropy loss ({\small NEG}, \cref{eq:neg_loss}).
Refer to the supplementary materials for the  hyperparameter $\lambda$ in \cref{eq:total_dtl_loss} used in our main experiments. For a fair comparison of unlearning methods, we adjusted $\lambda$ so that the differently unlearned models are compared with similar target accuracy.

\subsection{Piggyback learning accuracy of unlearned models}
\label{sec:placcuracy_unlearned}
\cref{tab:main_result} shows the comparison of four DTL models, which have similar target accuracy ($\tgtacc$). $\tgtacc$ is improved significantly in CIFAR-100 $\dtlarrow$ CIFAR-10-1\% experiment than {\small TGT} model, with negligible performance degradation compared to TL model. A similar trend is observed in CIFAR-100 $\dtlarrow$ STL-10-10\% case, where $\tgtacc$ is significantly higher than the {\small TGT} model with only a negligible penalty. Likewise, GC loss is superior to the others in TinyImageNet $\dtlarrow$ CIFAR-100 experiment.

The model with lower PL accuracy ($\placc$) successfully disposes of the source knowledge so is less susceptible to other piggyback tasks. Our experiments show that the proposed method (GC model) is the most effective in preventing piggyback learning.
Notably, when the pre-training task is CIFAR-100 and the piggyback task is CIFAR-100-10\%, the GC model successfully unlearns the source task knowledge by a significant gap in PL accuracy with other baselines.
The PL accuracy of the GC model showed a decrease of 24.53 percentage points compared to the TL model, whereas the PL accuracy of the best-performing baseline ({\small RAND}) was only 14.16 percentage points lower than that of the TL model.

In addition, when the SVHN-1\% dataset is piggybacked, only the GC model successfully prevents the full recovery of the PL accuracy of the TL model whereas others surpass the PL accuracy of the TL model. 
The proposed GC model significantly outperforms the baselines all while achieving target task performance comparable with the TL model across all benchmarks including the TinyImageNet experiments.

\begin{table}[t]
    \begin{center}
        \resizebox{\linewidth}{!}{
        \begin{tabular}{c c c c c c}
        \noalign{\smallskip}
        \toprule
        \multirow{2}{*}{Model}	&	\multicolumn{4}{c}{MIA strategy $\downarrow$}	&	\\
         \cmidrule(lr){2-5}
        	&	Adv. Dist	&	$\dag$Grad $w$	&	$\dag$Grad $x$	&	$\dag$WB	&	Avg.	\\
        \midrule
        TL	&	63.63	&	65.33	&	65.09	&	65.29	&	64.84	\\
        RAND	&	50.23	&	50.90	&	50.58	&	51.49	&	50.80	\\
        UNIF	&	50.40	&	51.10	&	50.64	&	53.14	&	51.32	\\
        NEG	&	\textbf{50.00}	&	51.91	&	50.74	&	56.93	&	52.40	\\
        GC	&	50.11	&	\textbf{50.69}	&	\textbf{50.45}	&	\textbf{50.52}	&	\textbf{50.44}	\\
        \bottomrule
        \noalign{\smallskip}
        \end{tabular}}
        \caption{Membership inference attack (MIA) accuracy on CIFAR-100 $\dtlarrow$ CIFAR-10 experiment. A lower value indicates better robustness to MIA, therefore more success in unlearning. MIA strategies used are \cite{AdvDistance, GradInvAttack, WB}. $\dag$ additionally involves training an attacker model.}
        \label{tab:MIA_summarized}
    \end{center}
\end{table}
\subsection{Robustness to membership inference attacks}
\label{sec:attack_result}
We investigated the potential security issue of membership information leakage on the private source data by MIA strategies, as mentioned in~\cref{sec:readout}. The success rate for white-box MIAs is reported in~\cref{tab:MIA_summarized} to evaluate the robustness of the DTL models. 
The result shows that all unlearned models are significantly more robust to MIAs than the TL model which has not been unlearned. Among them, the GC model is the most resistant to MIA attacks, demonstrating the lowest success rate across most attack strategies.
Notably, in the case of WB attack~\cite{WB} which utilizes intermediate features for fitting the attack model, the GC model remains considerably robust while other baselines are shown more vulnerable. It is because while other baselines focus on perturbing the output layer, GC loss directly perturbs the hidden layer representations through the gradient vectors, which results in significantly better resilience against multiple MIA-based privacy attacks. 

\subsection{Effectiveness of piggyback learning accuracy}
\label{sec:pla_effective}
We inspected the effectiveness of PL accuracy as an evaluation metric of knowledge disposal from two perspectives. First, our results show that the source accuracy of the DTL model cannot be used as a representative measure for estimating knowledge disposal. As shown in \cref{tab:reject_srcacc}, the {\small UNIF} model and {\small GC} model exhibit similar source accuracy ($\srcacc$), yet their PL accuracy on source data ($\placc$) differs significantly in both CIFAR-100$\dtlarrow$CIFAR-10-1\% and CIFAR-100$\dtlarrow$STL-10-10\% experiments. This is consistent with our observations that trivial unlearning can occur by simply choosing to disrupt the classification layer, resulting in degraded performance while the feature extractor remains intact.

\begin{figure}[t]
     \centering
     \begin{subfigure}{0.23\textwidth}
         \centering
         \includegraphics[width=\textwidth]{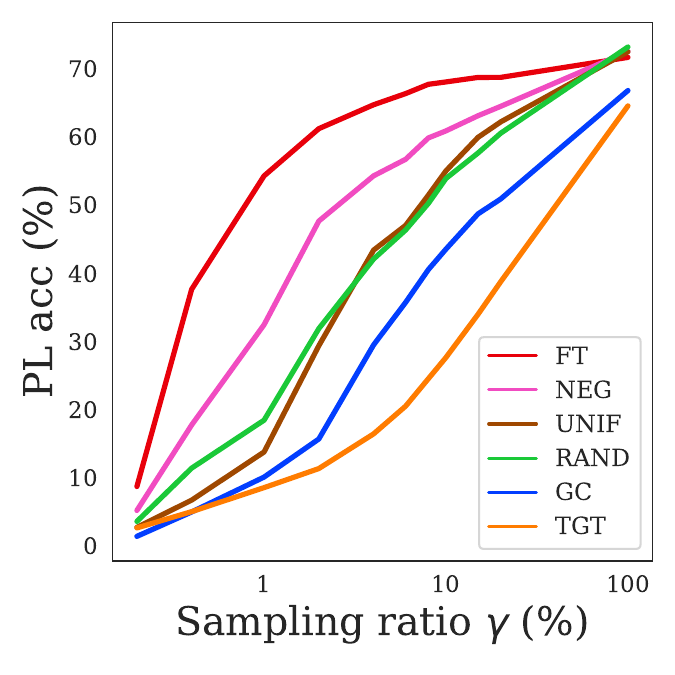}
         \caption{CIFAR-100-$\gamma$\%}
         \label{fig:nontarget_cif100}
     \end{subfigure}
     \begin{subfigure}{0.23\textwidth}
         \centering
         \includegraphics[width=\textwidth]{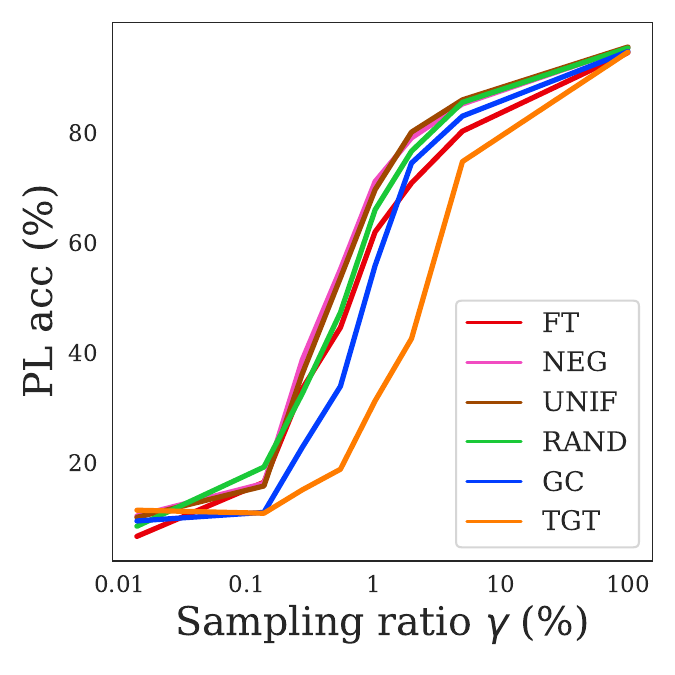}
         \caption{SVHN-$\gamma$\%}
         \label{fig:nontarget_svhn}
     \end{subfigure}
        \caption{PL accuracy in CIFAR-100$\dtlarrow$CIFAR-10-1\% experiment under varying dataset sampling rate $\gamma$.}
        \label{fig:nontarget_sampling}
\end{figure}

Furthermore, we have observed that PL accuracy stays an effective metric across an arbitrary size of the piggyback dataset. In \cref{fig:nontarget_sampling}, we examined the PL accuracy on the source data, CIFAR-100 (\cref{fig:nontarget_cif100}), and new downstream data, SVHN (\cref{fig:nontarget_svhn}), in CIFAR-100 $\dtlarrow$ CIFAR-10-1\% experiment. We simulated with an arbitrary size of the piggyback dataset by sampling $\gamma$\% of the original training data. 

\begin{table}[t]
    \centering
    \resizebox{\linewidth}{!}{
    \begin{tabular}{c | c c c | c c c}
    \noalign{\smallskip}
    \toprule
    \multirow{2}{*}{Model} & \multicolumn{3}{c|}{\scriptsize{CIFAR-100} $\dtlarrow$ \scriptsize{CIFAR-10-1\%}} & \multicolumn{3}{c}{\scriptsize{CIFAR-100} $\dtlarrow$ \scriptsize{STL-10-10\%}} \\ 
    \cmidrule{2-7}
    	&	$\srcacc$	&	$\tgtacc$	&	$\placc$	&	$\srcacc$	&	$\tgtacc$	&	$\placc$\\
    \midrule
    TL     	&	67.46	&	70.93	&	68.10	&	65.41	&	63.45	&	68.15\\
    RAND	&	1.64	&	69.00	&	53.94	&	2.02	&	62.47	&	56.88\\
    UNIF	&	2.74	&	71.41	&	55.06	&	7.03	&	63.04	&	55.67\\
    NEG	&	0.02	&	71.17	&	60.92	&	0.02	&	60.86	&	57.75\\
    GC	&	2.41	&	68.96	&	43.57	&	3.14	&	61.53	&	45.16\\
    \bottomrule
    \end{tabular}
    }
\caption{Comparison of source task accuracy ($\srcacc$) and PL accuracy ($\placc$) after DTL. $\srcacc$ does not serve as a proxy for estimating the degree of knowledge disposal  ($\placc$).}
\label{tab:reject_srcacc}
\end{table}

Interestingly, we found that the ranking of the PL accuracy on source data stays consistent across all ranges of sampling ratio (\cref{fig:nontarget_cif100}), which means that the data size has little effect on the validation of knowledge disposal. Measuring PL accuracy on another downstream task~(\cref{fig:nontarget_svhn}) has a similar effect on measuring the vulnerability of unlearned models on extra fine-tuning. In both experiments, the {\small TGT} model (orange) shows the weakest transfer for PL since it has not learned $\srcdata$. The GC model (blue) behaves relatively similarly to {\small TGT} while the TL model is much more susceptible to PL. This is because the TL does not forget the source task knowledge.
Note that when the sampling ratio $\gamma$ approaches 100\%, the PL accuracy metrics converge and become less discriminative. 

\subsection{Sensitivity analysis for $\lambda$}
\label{sec:exp_sensitivity}
\begin{figure}[t]
     \centering
     \begin{subfigure}{0.47\linewidth}
         \centering
         \includegraphics[width=\linewidth]{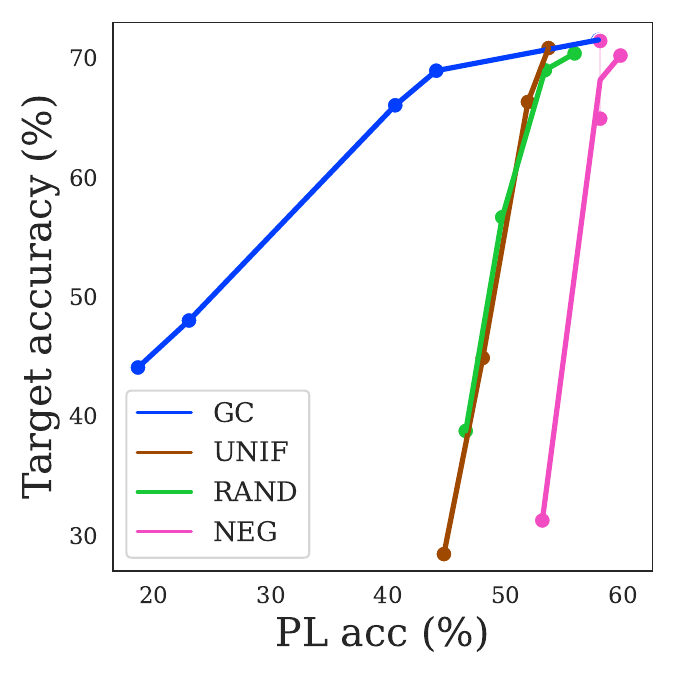}
         \caption{Unlearning loss}
         \label{fig:exp_unlearning_alpha}
     \end{subfigure}
     \begin{subfigure}{0.47\linewidth}
         \centering
         \includegraphics[width=\linewidth]{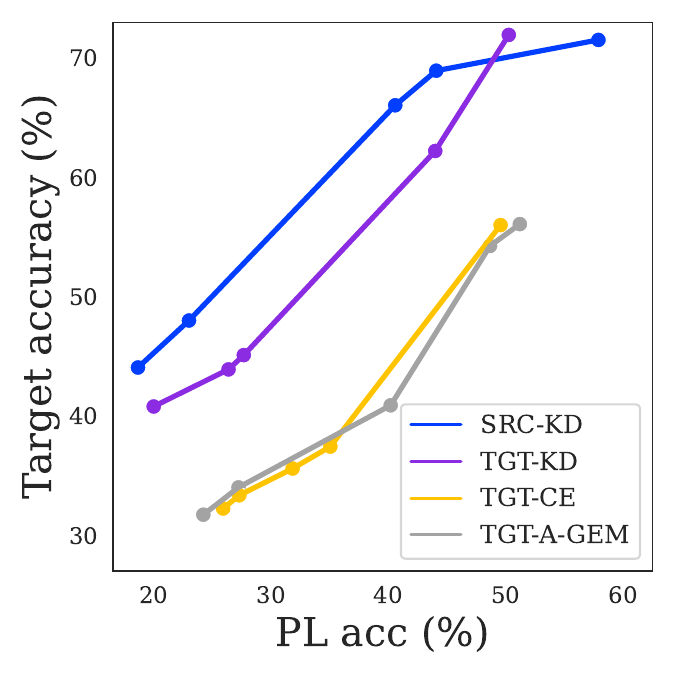}
         
         \caption{Knowledge retaining loss}
         \label{fig:exp_retaining_alpha}
     \end{subfigure}
    \caption{Trade-off of target accuracy and PL accuracy with varying $\lambda \in [0, 1]$. We experimented with different knowledge-retaining losses and unlearning losses.}
    \label{fig:exp_alpha}
\end{figure}

In this section, we discuss how the trade-off between model performance on the target accuracy and PL accuracy behaves across varying values of $\lambda$ in \cref{eq:total_dtl_loss} and characterizes the models. In \cref{fig:exp_alpha}, we show the PL accuracy vs. target accuracy curve by varying $\lambda$.
In ~\cref{fig:exp_unlearning_alpha}, we vary the unlearning losses while the KD loss is fixed as the source knowledge retaining loss. In ~\cref{fig:exp_retaining_alpha}, the unlearning loss is fixed to GC loss while we vary the knowledge retaining loss. Both experiments are conducted on the CIFAR-100 $\dtlarrow$ CIFAR-10-1\% experiment with CIFAR-100-10\% as the piggyback dataset.

Our results show that the powerful DTL performance of the combined loss of KD loss on the source data and GC loss is not a coincidence of a single value of $\lambda$, but it works for different levels of target and PL accuracy. In both experiments, the model which fits the unlearning goal of DTL is with high target accuracy and low PL accuracy, as it is plotted on the upper left region. As $\lambda$ increases, the target accuracy and PL accuracy decrease as the unlearning loss becomes dominant.
In ~\cref{fig:exp_unlearning_alpha}, it is observed that GC loss (blue) performs much better on the target task with the same level of PL accuracy. The GC loss retains target performance with significantly lower PL accuracy compared to the others which show catastrophically degraded target performance, even at a high PL accuracy.

Meanwhile, retaining the target task knowledge is another important factor in successful knowledge disposal. In \cref{fig:exp_retaining_alpha}, we compare different knowledge retaining losses. We compare the adopted KD loss with $\srcdata$ and the fine-tuned model for retaining the target knowledge ({\small SRC-KD}) as in \cref{eq:tgt_keeping_loss} against three other knowledge retaining baselines: 
KD loss with the target dataset ({\small TGT-KD}), training the model jointly with the typical cross-entropy loss with the target dataset ({\small TGT-CE}), cross-entropy loss replaced with A-GEM ({\small TGT-A-GEM})~\citep{GEM, A-GEM}, which is a constrained optimization technique used for continual learning. See Section C.1 in the supplementary materials for details on {\small TGT-A-GEM}.

We found that KD-based knowledge retaining losses ({\small SRC-KD, TGT-KD}) outperform the non-KD methods. They show significantly higher target accuracy than other baselines with the same level of PL accuracy. This is because KD not only retains the target knowledge but also transfers dark knowledge from the TL model. The results provide justifications for using {\small SRC-KD} over {\small TGT-KD} in knowledge retaining. As discussed in \cref{sec:dtl} and \cref{sec:retain_loss}, the amount of the target data is insufficient to represent the whole target distribution, whereas the larger source data facilitates better knowledge distillation from the TL model.

\section{Conclusion}
We propose a novel transfer learning scheme, named disposable transfer learning (DTL), which is designed to address the risk of piggybacking on a transfer-learned model when the model is released to the public. Our results highlight the promising potential of DTL towards preventing unauthorized exploitation of pre-trained weight for performance gain once the target task is adapted through transfer learning. To selectively dispose of the source knowledge, we propose a novel unlearning loss, coined gradient collision loss (GC loss). We have demonstrated that a combination of KD loss and GC loss successfully achieves DTL. 
Further, we propose an evaluation protocol named piggyback learning accuracy (PL accuracy) which verifies the susceptibility against piggyback learning. We have demonstrated that GC loss unlearns a model to make it less susceptible to malicious piggybacking through low PL accuracy, emphasizing the effectiveness of our method.

\subsection*{Limitations and Future Works} 
Our work focused on establishing the DTL paradigm and its implementation. While our GC loss has shown superiority compared to the basic methods, there remains considerable room for improving the unlearning objective. Specifically, better integration of the knowledge retaining and unlearning objectives needs to be further explored; in our approach, we chose to simply optimize the sum of two separate objectives. Additionally, our studies were confined to relatively small-to-medium-scale datasets and model architecture. We believe that our work provides a foundation for future research for DTL. We encourage subsequent studies to design a better-integrated objective and investigate DTL within larger, more realistic contexts.

\subsection*{Acknowledgements} 
This work was supported by Institute of Information \& communications Technology Planning \& Evaluation (IITP) grant funded by the Korea government (MSIT) (No. 2022-0-00951, Development of Uncertainty-Aware Agents Learning by Asking Questions).

{\small
\bibliographystyle{ieee_fullname}
\bibliography{egbib}
}

\clearpage

\appendix
\section{Experimental configurations}

In this paper, we set different learning rates, the number of epochs, batch sizes, and the number of GPU processes for each training step.

\textbf{Optimization}\quad We used cross-entropy loss for learning. For optimization, we used the SGD optimizer with \texttt{weight\_decay=1e-4} and \texttt{momentum=0.9}. The learning rate is scheduled by a cosine annealing scheme with the initial learning rate $\eta_0$ described in the following paragraph.

\textbf{Hyperparameters}\quad In pre-training of TL and knowledge disposal, we set the initial learning rate to $\eta_0=0.05$ for training 90 epochs with 4 GPUs. Because those stages are trained with the full size of the source dataset, it is acceptable to use larger batch size and more GPUs for faster training. Especially, for GC loss, we divide each mini-batch into $c=4$ chunks so that each GPU calculates for a chunk. In fine-tuning of TL and piggyback learning (PL), the model is trained with an initial learning rate of $\eta_0=0.01$ for training 30 epochs with two GPUs. We reduced the batch size because the scale of training data is much smaller, but we matched the number of batches per GPU to 32.

For the TinyImageNet experiments, the dataset was downsampled to a resolution of $36\times36$ and random-cropped to $32\times32$ to be consistent with the input layer used for the CIFAR dataset, our target task. For pre-training source task, we used a batch size of 64, running on two GPUs, with an initial learning rate of 0.05 and trained over 90 epochs. For training target task, we employed a batch size of 32 across two GPUs with an initial learning rate of 0.01 and trained over 30 epochs. For the knowledge disposal stage, the settings were consistent with the experiments using CIFAR-100 as the source task: a batch size of 128, utilizing four GPUs, with an initial learning rate of 0.05. Concerning piggyback learning, the TinyImageNet piggyback was trained using the same setting as the knowledge disposal stage, whereas other datasets were piggybacked with a batch size of 128, using two GPUs, over 30 epochs, with a learning rate of 0.01.

\textbf{Classification layers}\quad We assign classification layers per task which are linear layers that take the extracted features from the shared feature network, so the classification layer is independent between the preceding and the proceeding tasks.

\section{Details on the baseline methods}

In the case of RAND, UNIF, NEG, and GC models, we set the best model with different values of $\lambda$ for each unlearning loss in each experiment as in  \cref{tab:unlearning_lambda}.

In addition, two additional baselines are reported in this material. The PRE and TGT model behaves similarly to the TL and TGT model, respectively. First, we estimated the PRE model, the output model of pre-training in TL. The PRE model is estimated to analyze the effect of pre-training and to compare the performance with the TL model. Likewise, we analyzed the SCR model, which is a randomly initialized model without any training. The TGT model is more suitable for comparing the performance gain with other unlearned models than the SCR model because the TGT model has sufficiently low target performance and it has little target knowledge.

\begin{table}[t]
\begin{center}
\resizebox{\linewidth}{!}{
\begin{tabular}{c c c c c }
\noalign{\smallskip}
\toprule
Experiment                          &   RAND    &   UNIF    &   NEG     &   GC \\
\midrule
CIFAR-100$\dtlarrow$CIFAR-10-1\%      &   0.9000     &   0.9000     &   0.9750   &   0.3000     \\
CIFAR-100$\dtlarrow$STL-10-10\%        &   0.6000     &   0.8500    &   0.9900    &   0.2600    \\
TinyImageNet$\dtlarrow$CIFAR-100-10\%        &   0.8750     &   0.9700    &   0.9875    &   0.2600    \\
\bottomrule
\end{tabular}}
\caption{$\lambda$ values used for the experiments.}
\label{tab:unlearning_lambda}
\end{center}
\end{table}

\section{Discussions on the DTL objective functions}
\subsection{Modified A-GEM for knowledge retaining}
\label{sec:gem_forgetting}

In \cref{sec:exp_sensitivity} and \cref{fig:exp_retaining_alpha}, we compared the DTL performance of various retaining losses. Among them, TGT-A-GEM is a modified knowledge-retaining method based on the A-GEM method \cite{GEM, A-GEM} for continual learning. It prevents catastrophic forgetting of previous tasks by updating the gradient of the current task in a direction so as not to increase the loss of the previous tasks.

Drawing inspiration from the mechanism, we adopt the A-GEM algorithm to effectively retain the target knowledge. To prevent catastrophic forgetting of the target task, we added the constraint to our optimizing goal, \cref{eq:total_dtl_loss}, as below. $\theta^t$ refers to the parameter state after $t$-th gradient descent update.
\begin{align}
    \textit{minimize}  \quad & \ldtl(\theta^t) \label{eq:optimizing}\\
    \textit{subject to}  \quad & \llearn(\theta^t) \leq {\llearn(\theta^{t-1})} \label{eq:optimizing_const}
\end{align}
\cref{eq:optimizing_const} indicates the knowledge retaining loss has to be non-increasing as $\theta^t$ is updated by SGD, which implies learning the target data $\tgtdata$ not to be restrained by unlearning the source data $\srcdata$.
The difference between the original A-GEM algorithm and our TGT-A-GEM is the object of knowledge retaining and the kind of minimizing objective. In continual learning, the loss on previous tasks has to be non-increasing, but in our study, the loss on retaining data has to be non-increasing. Also, we minimize $\loss_{dtl}$, a linearly interpolated loss, but A-GEM minimizes a single kind of loss calculated from the current task.

From now on, we follow Equation (4) in \cite{GEM} to obtain the gradient update formula. We rephrase \cref{eq:optimizing,eq:optimizing_const} with respect to $\nabla\loss$. 
\begin{align}
\label{eq:grad_opt}
    \text{minimize}_{\nabla\tilde{\loss}} \quad & \frac{1}{2} \lVert \gdtl - \nabla\tilde{\loss}\rVert^2_2\\
    \text{subject to} \quad & {\nabla\tilde{\loss}}^\top \glearn \geq 0 \label{eq:optimizing_const_grad}
\end{align}
According to \cref{eq:optimizing_const_grad}, if ${\nabla\tilde{\loss}}^\top \glearn$ is non-negative, then $\nabla\tilde{\loss} = \gdtl$. If the inner product is negative, $\nabla\tilde\loss$ is projected to $\glearn$.
The policy for gradient updating reflects that the gradient of DTL loss should be updated not to disturb the learning of the target task if there is a directional conflict between the learning gradient and the unlearning gradient. Finally, our solution for that problem is as  \cref{eq:projection}.
\begin{equation}
    \label{eq:projection}
    \resizebox{0.91\hsize}{!}{$
    \nabla\tilde{\loss} = 
    \begin{cases}
        \gdtl, &\mbox{if }  \gdtl^\top \glearn \geq 0 \\
        \gdtl - \frac{\gdtl^\top \glearn}{\glearn^\top \glearn}\glearn, &\mbox{otherwise}
    \end{cases}
    $}
\end{equation}

We adopted the novel gradient update policy for DTL, but as shown in \cref{fig:exp_retaining_alpha}, it is found that the knowledge retaining performance of A-GEM on target data (TGT-A-GEM) is not clearly distinguished from naive knowledge retaining with the cross-entropy loss with the target data (TGT-CE). Fundamentally, due to the small scale of the target data, the models with retaining knowledge from the TL model by distillation outperform others.

\subsection{Normalized gradient collision loss}

We conduct an extra investigation on a variant of the GC loss where we eliminate the magnitude information from the gradient, named Normalized Gradient Collision (NGC) loss.
\begin{definition}[Normalized Gradient Collision loss]
\label{def:ngc}
Normalized GC loss is a variant of the GC loss (\cref{eq:gc_loss_ideal}). It focuses on the angle between the loss and ignores the scale of gradients by minimizing only the cosine similarity of the gradient pairs.
\begin{equation}
    \label{eq:normalized_grad_collision}
    \loss_{ngc}(\dset, \theta) = {\frac{1} {\binom{c} {2}}}{\sum_{m \neq n} \frac{\nabla \ell_m(\theta)^\top \nabla \ell_n(\theta)}{\lVert\nabla \ell_m(\theta)\rVert\lVert \nabla \ell_n(\theta)\rVert}}
\end{equation}
\end{definition}
In practice, we have found that regularizing the grad norm as well as minimizing the variance, which corresponds to the GC model, results in better performance. We show an analysis of the NGC and GC loss in \cref{sec:ngc_freeze}.

 \begin{table*}[t]
\begin{center}
\begin{subtable}[c]{\linewidth}
\centering
\begin{tabular}{c c c c c c }
\noalign{\smallskip}
\toprule
\multirow{2}{*}{Model} &	\multirow{2}{*}{$\srcacc$}	&	\multirow{2}{*}{$\tgtacc$$\uparrow$}	&	\multicolumn{3}{c}{PL accuracy $\downarrow$} \\
\cmidrule{4-6}
&   &   &   CIFAR-100-10\% & STL-10-10\%	&	SVHN-1\% \\
\midrule
SCR		&	1.00	&	8.53	&	25.75	&	39.58	&	22.40		\\
PRE		&	71.54	&	11.41	&	68.20	&	63.45	&	60.82		\\
TL		&	67.46	&	70.93	&	68.10	&	65.25	&	61.97		\\
TGT		&	1.22	&	35.12	&	27.67	&	41.90	&	31.33		\\
\midrule
RAND	&	1.64	&	69.00	&	53.94	&	62.25	&	65.98		\\
UNIF	&	2.74	&	71.41	&	55.06	&	64.08	&	69.63		\\
NEG		&	0.02	&	71.17	&	60.92	&	65.16	&	71.10		\\
GC		&	2.41	&	68.96	&	43.57	&	57.78	&	55.87		\\
\bottomrule
\noalign{\smallskip}
\end{tabular}
\caption{CIFAR-100$\dtlarrow$CIFAR-10-1\%}
\end{subtable}

\begin{subtable}[c]{\linewidth}
\centering
\begin{tabular}{c c c c c c}
\noalign{\smallskip}
\toprule
\multirow{2}{*}{Model} &		\multirow{2}{*}{$\srcacc$}	&	\multirow{2}{*}{$\tgtacc$$\uparrow$}	&	\multicolumn{3}{c}{PL accuracy $\downarrow$}\\
\cmidrule{4-6}
& &&CIFAR-100-10\%& CIFAR-10-1\%	&	SVHN-1\%\\
\midrule
SCR		&	1.09	&	10.00	&	25.75	&	35.12	&	22.40	\\
PRE		&	71.54	&	10.56	&	68.20	&	70.93	&	60.82	\\
TL		&	65.41	&	63.45	&	68.15	&	72.12	&	60.82	\\
TGT		&	1.15	&	39.58	&	27.76	&	36.94	&	33.85	\\
\midrule
RAND	&	2.02	&	62.47	&	56.88	&	70.21	&	67.24	\\
UNIF	&	7.03	&	63.04	&	55.67	&	69.57	&	67.22	\\
NEG		&	0.02	&	60.86	&	57.75	&	67.97	&	70.47	\\
GC		&	3.14	&	61.53	&	45.16	&	61.19	&	56.63	\\
\bottomrule
\noalign{\smallskip}
\end{tabular}
\caption{CIFAR-100$\dtlarrow$STL-10-10\%}
\end{subtable}

\begin{subtable}[c]{\linewidth}
\centering
\begin{tabular}{c c c c c c}
\noalign{\smallskip}
\toprule
\multirow{2}{*}{Model} &		\multirow{2}{*}{$\srcacc$}	&	\multirow{2}{*}{$\tgtacc$$\uparrow$}	&	\multicolumn{3}{c}{PL accuracy $\downarrow$}\\
\cmidrule{4-6}
& &&TinyImageNet-6\%& STL-10-10\%	&	CIFAR-10-1\%\\
\midrule
SCR 	&	0.46	&	1.04	&	14.90	&	35.91	&	30.72 \\
PRE	    &	54.87	&	0.36	&	38.95 	&	69.82   &	75.15\\
FT      &	38.06	&	55.99	&	39.12	&	67.71	&   72.08  \\
TGT     &   0.41    &   25.16   &   17.16   &   45.44   &   40.82   \\
\midrule
RAND	&	1.22	&	53.79	&	29.20	&	65.81	&   67.40   \\
UNIF	&	5.66	&	54.79	&	28.00	&	64.21	&	68.60 \\
NEG     &	0.06	&	53.96	&	27.20	&	64.94	&	68.21  \\
GC      &	5.46	&	54.35	&	26.40	&	62.83	&	64.18  \\
\bottomrule
\noalign{\smallskip}
\end{tabular}
\caption{TinyImageNet$\dtlarrow$CIFAR-100-10\%}
\end{subtable}

\caption{The transition of source and target accuracy for each DTL stage and piggyback learning in CIFAR-100$\dtlarrow$CIFAR-10-1\%, CIFAR-100$\dtlarrow$STL-10-10\%, and TinyImageNet$\dtlarrow$CIFAR-100-10\% experiments.}
\label{tab:naive_main_result}
\end{center}
\end{table*}

\subsection{KL-divergence cannot represent unlearning}
\label{sec:kldiv_no_forgetting}
In this section, we further discuss why simply minimizing the log-likelihood cannot lead to unlearning through a counter-example. In the main manuscript, we have defined the likelihood-minimizing objective function as $\mathcal{L}_{negative}$ in \cref{eq:neg_loss}. 

The KL-divergence between a target distribution $P$ and model distribution $Q$ is as follows:
\begin{equation}
    \kldiv{P(y|x)}{Q(y|x)} = \sum_{y\in\mathcal{Y}} P(y|x)\text{log}\frac{P(y|x)}{Q(y|x)}.
\end{equation}
It is seen that $D_{KL}(P||Q)$ diverges if there exists a sample $\hat y \in \mathcal{Y}$ such that $Q(y|x) \to 0$ and $P(y|x) > 0$ {\cite{Murphy_VI, intro_vi}}. Unlike the minimization of KL-divergence which results in distributional similarity, KL-divergence can be trivially maximized by degenerating the softmax score of a single sample.
Moreover, a trivial maximization can be achieved by perturbing a few neurons on the uppermost layers, which is highly related to class prediction value, so it is much easier to increase the KL divergence with the intact feature extractor.

\section{Extended experimental results}
\subsection{Full result of the main experiments}

In \cref{tab:naive_main_result}, we provide the raw results of our CIFAR-100$\dtlarrow$CIFAR-10-1\%, CIFAR-100$\dtlarrow$STL-10-10\%, and TinyImageNet$\dtlarrow$CIFAR-100-10\% experiment, including source accuracy, target accuracy, and PL accuracy. It is visualized in various ways for motivating our work in \cref{fig:finetuning_over_finetuning}, or highlighting the relative performance change as shown in \cref{tab:main_result} and \cref{tab:reject_srcacc}.  From now on, we give a straightforward explanation of how we interpret and focus on the meaningful result based on the table.

\begin{figure}[t]
     \centering
     \begin{subfigure}{0.48\linewidth}
          \centering
          \includegraphics[width=0.95\linewidth]{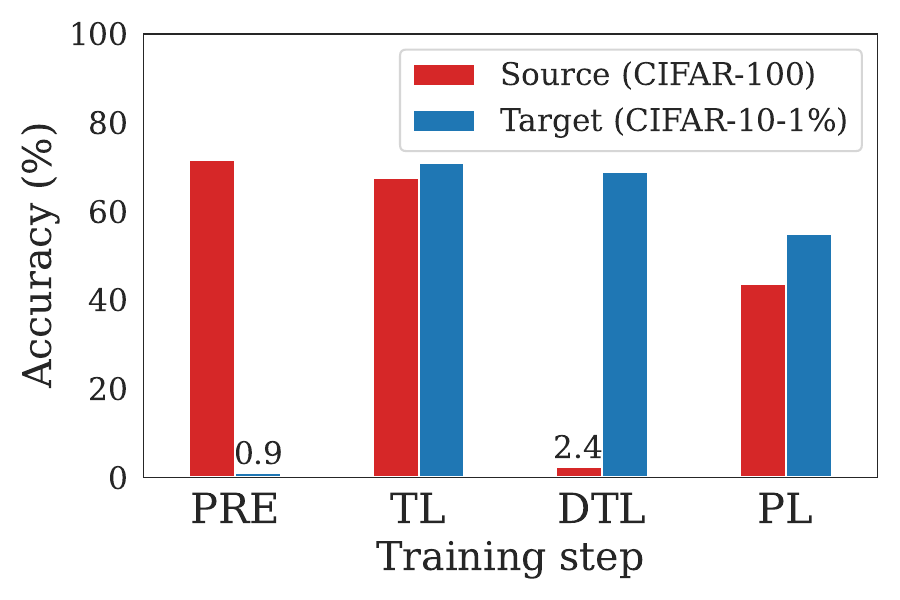}
          \caption{\scriptsize CIFAR-100$\dtlarrow$CIFAR-10-1\%}
          \label{fig:transition_cif10}
     \end{subfigure}
     \begin{subfigure}{0.48\linewidth}
          \centering
          \includegraphics[width=0.95\linewidth]{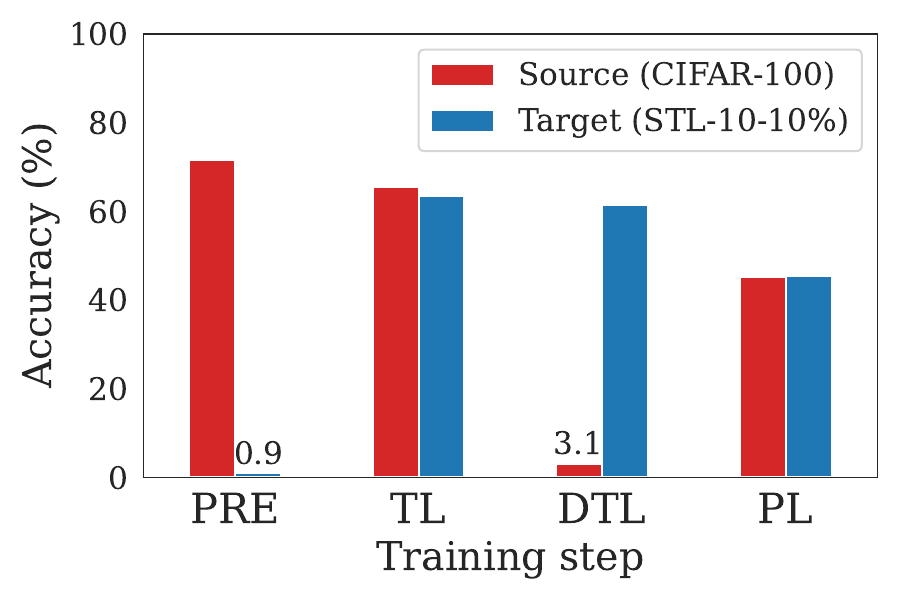}
          \caption{\scriptsize CIFAR-100$\dtlarrow$STL-10-10\%}
          \label{fig:transition_stl10}
     \end{subfigure}
        \caption{Performance transition for each training step. The model is repeatedly fine-tuned from the SCR model in order of pre-training, fine-tuning, knowledge disposal, and piggyback learning. In other words, we can get the output models in the sequence of SCR $\rightarrow$ PRE $\rightarrow$ TL $\rightarrow$ DTL $\rightarrow$ PL.}
        \label{fig:transition}
\end{figure}

The performance of the fine-tuning on different initializers, as plotted in \cref{fig:finetuning_over_finetuning}, is based on the PL accuracies of SCR (red), PRE(green), and TL (green) model. It motivates us to propose DTL, as claimed in the Introduction section, that the fine-tuned model also is a competitive representation model as the pre-trained model.

In \cref{sec:placcuracy_unlearned}, we emphasize the performance gain and penalty on the target task by reporting the relative performance in \cref{tab:main_result}. It corresponds to the performance gap with TGT / TL models and our unlearned models in \cref{tab:naive_main_result}. PL accuracy is reported with the amount of performance changed from the TL model to our unlearned models. Comparing the relative performance of the GC model is a simpler way to evaluate its powerful DTL performance than comparing its absolute performance.

\begin{figure}[t]
     \centering
     \begin{subfigure}{0.23\textwidth}
          \centering
          \includegraphics[width=0.95\linewidth]{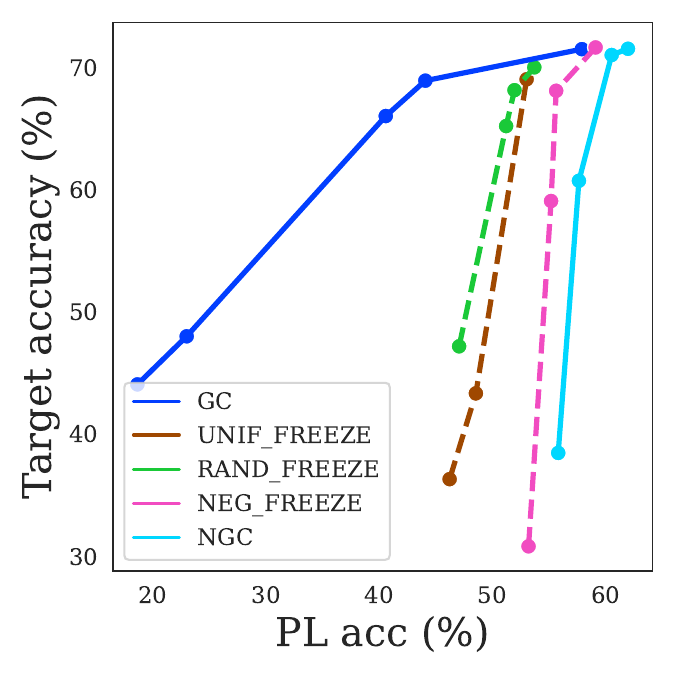}
          \caption{CIFAR-100$\dtlarrow$CIFAR-10-1\% with unlearning variants}
          \label{fig:ngc_freeze}
     \end{subfigure}
     \begin{subfigure}{0.23\textwidth}
          \centering
          \includegraphics[width=0.95\linewidth]{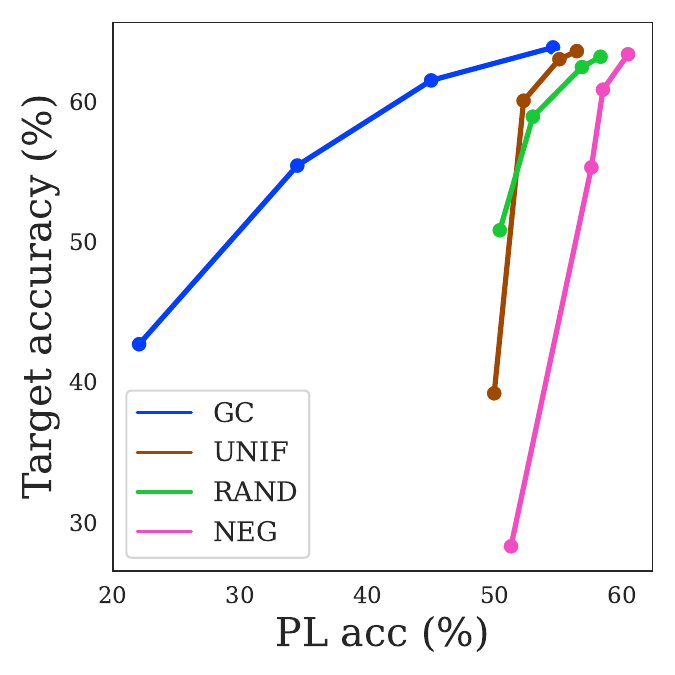}
          \caption{CIFAR-100$\dtlarrow$STL-10-10\% experiment}
          \label{fig:alpha_tuning_stl}
     \end{subfigure}
        \caption{(a) The result of the variants of unlearning losses in CIFAR-100$\dtlarrow$CIFAR-10-1\% experiment, which is plotted in the same manner with \cref{fig:exp_unlearning_alpha}. NGC loss model behavior is significantly different from GC loss model. Furthermore, baseline losses with frozen FC behave similarly as reported in the main manuscript. (b) The result of the same analysis with \cref{fig:exp_unlearning_alpha} in CIFAR-100$\dtlarrow$STL-10-10\% experiment.}
        \label{fig:further_alpha_tuning}
\end{figure}

Also, we show the appropriateness of PL accuracy by reporting the absolute value of source accuracy and PL accuracy in  \cref{tab:reject_srcacc}. We mainly focused on the irrelevance of performance degradation (source accuracy) and knowledge disposal (PL accuracy). Note that the reported PL accuracy is the validation accuracy after training with a portion of the source data, CIFAR-100.

\subsection{Variants of unlearning losses}
\label{sec:ngc_freeze}
As well as the unlearning baselines in the main manuscript, we also explore the variants in this section. NGC model, defined at \cref{eq:normalized_grad_collision}, and the variants of unlearning baselines are compared against our GC model in \cref{fig:ngc_freeze}. The unlearning baselines, which are plotted in dashed lines, are unlearned by freezing the last classification layer of the source task and applying corresponding fooling losses. We freeze the classifier layer to test whether the baseline unlearning losses effects only the FC layer or not.

Among the models in \cref{fig:ngc_freeze}, we observed that the GC loss performs the best. The frozen variants (dashed lines) behave similarly to the result in \cref{fig:exp_unlearning_alpha}. This implies that the baseline fooling losses only affect a few uppermost non-frozen layers.
Interestingly, we have observed that the NGC loss performs worse than all other baselines unlearning loss, which shows that penalizing the gradient norm is also an important factor in gradient collision unlearning.

\begin{figure}[t]
     \centering
     \begin{subfigure}{0.23\textwidth}
         \centering
         \includegraphics[width=\linewidth]{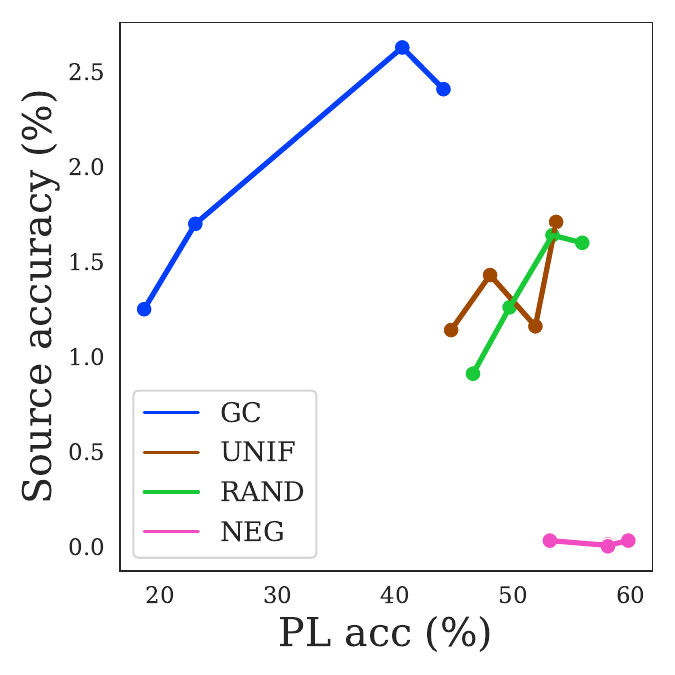}
         \caption{Source accuracy vs PL accuracy. $\rho_s=-0.505$.}
         \label{fig:src_pla}
     \end{subfigure}
     \begin{subfigure}{0.23\textwidth}
         \centering
         \includegraphics[width=\linewidth]{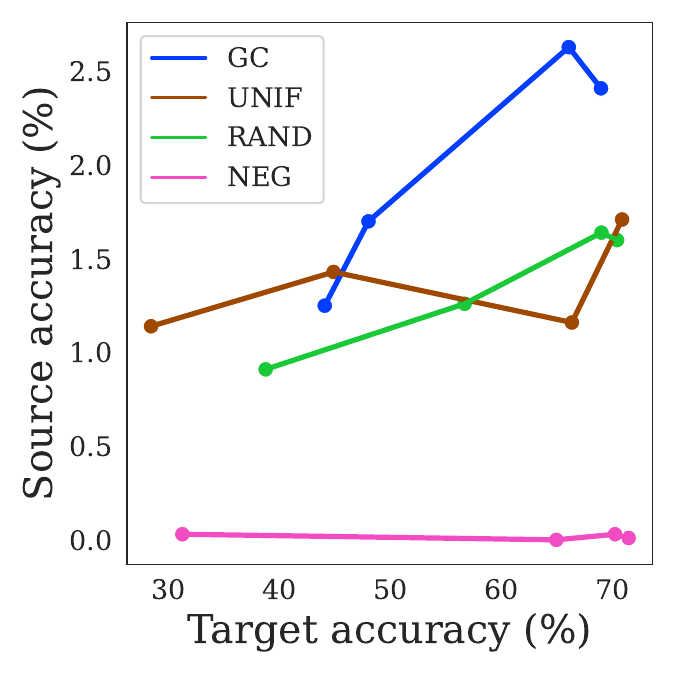}
         \caption{Source accuracy vs Target accuracy. $\rho_s=0.199$.}
         \label{fig:tgt_src}
     \end{subfigure}
        \caption{The trade-off relationship of PL acc vs source accuracy and target accuracy vs source accuracy in CIFAR-100$\dtlarrow$CIFAR-10-1\% experiment with varying $\lambda$ for each unlearning loss. $\rho_s$ indicates the Spearman correlation coefficient.}
        \label{fig:stc_tgt}
\end{figure}
\begin{table*}[t]
\begin{center}
\resizebox{\textwidth}{!}{
\begin{tabular}{c | c  c | c  c | c c |cc}
\noalign{\smallskip}
\hline
\multirow{2}{*}{MIA strategy} &	\multicolumn{2}{c|}{SCRATCH}&  \multicolumn{2}{c|}{PRE}&	\multicolumn{2}{c|}{TL} & \multicolumn{2}{c}{TGT}\\	
	&		AUROC	&	Accuracy	&   AUROC	&	Accuracy	&	AUROC	&	Accuracy	&	AUROC	&	Accuracy	\\
 \hline
Softmax	&	49.61	$_{\pm	0.23	}$&	50.24	$_{\pm	0.13	}$&	67.96	$_{\pm	0.23	}$&	66.09	$_{\pm	0.25	}$&	63.71	$_{\pm	0.24	}$&	61.68	$_{\pm	0.28	}$&	49.86	$_{\pm	0.29	}$&	50.42	$_{\pm	0.20	}$\\
Mentr.	&	50.06	$_{\pm	0.25	}$&	50.42	$_{\pm	0.17	}$&	70.14	$_{\pm	0.20	}$&	68.49	$_{\pm	0.19	}$&	67.33	$_{\pm	0.20	}$&	65.15	$_{\pm	0.18	}$&	50.12	$_{\pm	0.19	}$&	50.44	$_{\pm	0.13	}$\\
Loss	&	50.00	$_{\pm	0.05	}$&	50.06	$_{\pm	0.04	}$&	69.93	$_{\pm	0.20	}$&	68.50	$_{\pm	0.19	}$&	67.25	$_{\pm	0.20	}$&	65.19	$_{\pm	0.18	}$&	49.97	$_{\pm	0.04	}$&	50.01	$_{\pm	0.00	}$\\
Grad Norm	&	50.04	$_{\pm	0.25	}$&	50.40	$_{\pm	0.16	}$&	69.98	$_{\pm	0.20	}$&	68.60	$_{\pm	0.19	}$&	66.83	$_{\pm	0.21	}$&	65.16	$_{\pm	0.18	}$&	50.12	$_{\pm	0.19	}$&	50.44	$_{\pm	0.12	}$\\
Adv. Dist	&	50.34	$_{\pm	0.21	}$&	50.66	$_{\pm	0.11	}$&	63.73	$_{\pm	0.18	}$&	64.34	$_{\pm	0.05	}$&	63.03	$_{\pm	0.19	}$&	63.63	$_{\pm	0.14	}$&	50.36	$_{\pm	0.23	}$&	50.70	$_{\pm	0.17	}$\\

$\dag$Grad $w$	&	49.91	$_{\pm	0.39	}$&	50.46	$_{\pm	0.26	}$&	70.66	$_{\pm	0.52	}$&	68.75	$_{\pm	0.31	}$&	67.55	$_{\pm	0.46	}$&	65.33	$_{\pm	0.25	}$&	49.74	$_{\pm	0.34	}$&	50.41	$_{\pm	0.16	}$\\
$\dag$Grad $x$	&	49.96	$_{\pm	0.39	}$&	50.55	$_{\pm	0.20	}$&	71.00	$_{\pm	0.43	}$&	68.72	$_{\pm	0.30	}$&	67.48	$_{\pm	0.35	}$&	65.09	$_{\pm	0.24	}$&	50.15	$_{\pm	0.44	}$&	50.67	$_{\pm	0.25	}$\\
$\dag$Int. Outs	&	49.80	$_{\pm	0.45	}$&	50.46	$_{\pm	0.27	}$&	52.17	$_{\pm	0.39	}$&	51.93	$_{\pm	0.36	}$&	51.67	$_{\pm	0.53	}$&	51.65	$_{\pm	0.42	}$&	49.88	$_{\pm	0.33	}$&	50.48	$_{\pm	0.20	}$\\
$\dag$WB	&	49.87	$_{\pm	0.27	}$&	50.46	$_{\pm	0.16	}$&	70.82	$_{\pm	0.43	}$&	68.60	$_{\pm	0.30	}$&	67.88	$_{\pm	0.38	}$&	65.29	$_{\pm	0.26	}$&	50.08	$_{\pm	0.48	}$&	50.59	$_{\pm	0.33	}$\\

\hline
\hline
\multirow{2}{*}{MIA strategy}&	\multicolumn{2}{c|}{GC}& \multicolumn{2}{c|}{RAND}& \multicolumn{2}{c|}{NEG}& \multicolumn{2}{c}{UNIF}\\	
	&		AUROC	&	Accuracy	&	AUROC	&	Accuracy	&	AUROC	&	Accuracy &	AUROC	&	Accuracy	\\
 \hline
Softmax	&	50.76	$_{\pm	0.22	}$&	50.93	$_{\pm	0.16	}$&	51.51	$_{\pm	0.32	}$&	51.47	$_{\pm	0.22	}$&	50.91	$_{\pm	0.26	}$&	50.96	$_{\pm	0.22	}$&	50.65	$_{\pm	0.28	}$&	50.79	$_{\pm	0.21	}$\\
Mentr.	&	50.23	$_{\pm	0.29	}$&	50.47	$_{\pm	0.19	}$&	51.41	$_{\pm	0.19	}$&	51.41	$_{\pm	0.14	}$&	52.06	$_{\pm	0.26	}$&	51.90	$_{\pm	0.19	}$&	53.88	$_{\pm	0.13	}$&	53.18	$_{\pm	0.11	}$\\
Loss	&	50.23	$_{\pm	0.29	}$&	50.47	$_{\pm	0.19	}$&	51.42	$_{\pm	0.19	}$&	51.42	$_{\pm	0.14	}$&	56.46	$_{\pm	0.24	}$&	55.01	$_{\pm	0.18	}$&	53.88	$_{\pm	0.13	}$&	53.18	$_{\pm	0.11	}$\\
Grad Norm	&	49.90	$_{\pm	0.29	}$&	50.34	$_{\pm	0.14	}$&	49.49	$_{\pm	0.23	}$&	50.20	$_{\pm	0.07	}$&	52.65	$_{\pm	0.22	}$&	52.43	$_{\pm	0.24	}$&	49.23	$_{\pm	0.22	}$&	50.29	$_{\pm	0.09	}$\\
Adv. Dist	&	50.05	$_{\pm	0.06	}$&	50.11	$_{\pm	0.05	}$&	50.22	$_{\pm	0.06	}$&	50.23	$_{\pm	0.06	}$&	49.99	$_{\pm	0.00	}$&	50.00	$_{\pm	0.00	}$&	50.39	$_{\pm	0.07	}$&	50.40	$_{\pm	0.08	}$\\
$\dag$Grad $w$	&	50.23	$_{\pm	0.47	}$&	50.69	$_{\pm	0.29	}$&	50.63	$_{\pm	0.42	}$&	50.90	$_{\pm	0.31	}$&	52.07	$_{\pm	0.54	}$&	51.91	$_{\pm	0.41	}$&	50.83	$_{\pm	0.48	}$&	51.10	$_{\pm	0.35	}$\\
$\dag$Grad $x$	&	49.84	$_{\pm	0.51	}$&	50.45	$_{\pm	0.29	}$&	50.12	$_{\pm	0.47	}$&	50.58	$_{\pm	0.20	}$&	50.40	$_{\pm	0.55	}$&	50.74	$_{\pm	0.30	}$&	50.21	$_{\pm	0.60	}$&	50.64	$_{\pm	0.38	}$\\
$\dag$Int. Outs	&	49.89	$_{\pm	0.59	}$&	50.52	$_{\pm	0.31	}$&	50.15	$_{\pm	0.62	}$&	50.71	$_{\pm	0.38	}$&	50.86	$_{\pm	0.62	}$&	51.10	$_{\pm	0.42	}$&	49.90	$_{\pm	0.51	}$&	50.52	$_{\pm	0.28	}$\\
$\dag$WB	&	49.99	$_{\pm	0.42	}$&	50.52	$_{\pm	0.23	}$&	51.46	$_{\pm	0.64	}$&	51.49	$_{\pm	0.42	}$&	59.17	$_{\pm	0.51	}$&	56.93	$_{\pm	0.39	}$&	53.59	$_{\pm	0.52	}$&	53.14	$_{\pm	0.40	}$\\
\hline
\noalign{\smallskip}
\end{tabular}
}
\caption{The success rate of MIAs of models in CIFAR-100$\rightarrow$CIFAR-10-1\%. MIA strategies used are \cite{AdvDistance, GradInvAttack, WB}. $\dag$ involves training an attacker model.}
\label{tab:mia_full}
\end{center}
\end{table*}

\begin{figure*}[ht]
     \centering
     \begin{subfigure}{0.23\textwidth}
  \centering
  \includegraphics[width=0.95\linewidth]{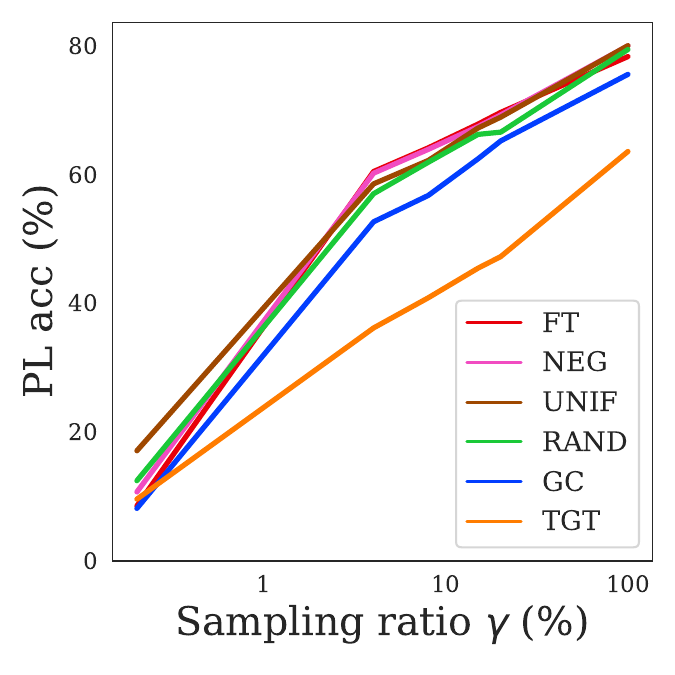}
  \caption{\footnotesize STL10 PL accuracy in \\CIFAR-100$\dtlarrow$CIFAR-10-1\%.}
  \label{fig:pla_cif10_stl10}
     \end{subfigure}
     \begin{subfigure}{0.23\textwidth}
  \centering
  \includegraphics[width=0.95\linewidth]{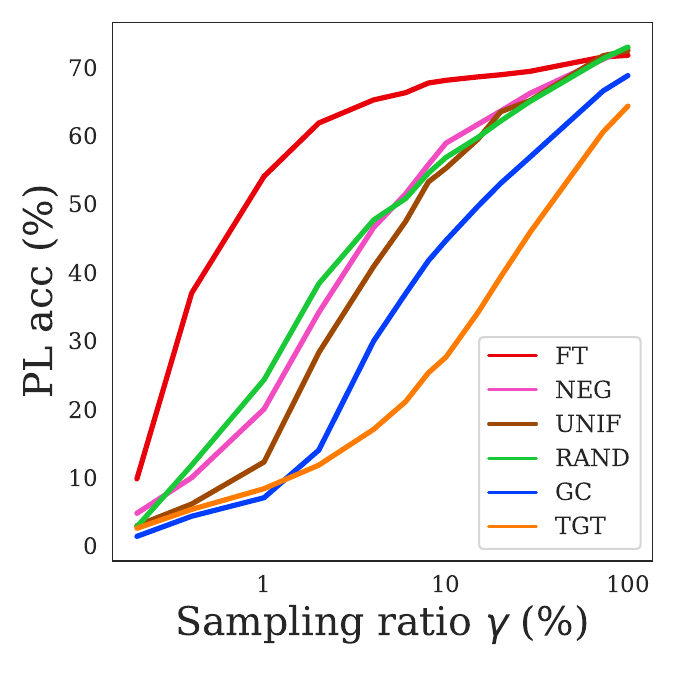}
  \caption{\footnotesize CIFAR-100 PL accuracy in \\CIFAR-100$\dtlarrow$STL-10-10\%.}
  \label{fig:pla_stl10_cif100}
     \end{subfigure}
          \begin{subfigure}{0.23\textwidth}
  \centering
  \includegraphics[width=0.95\linewidth]{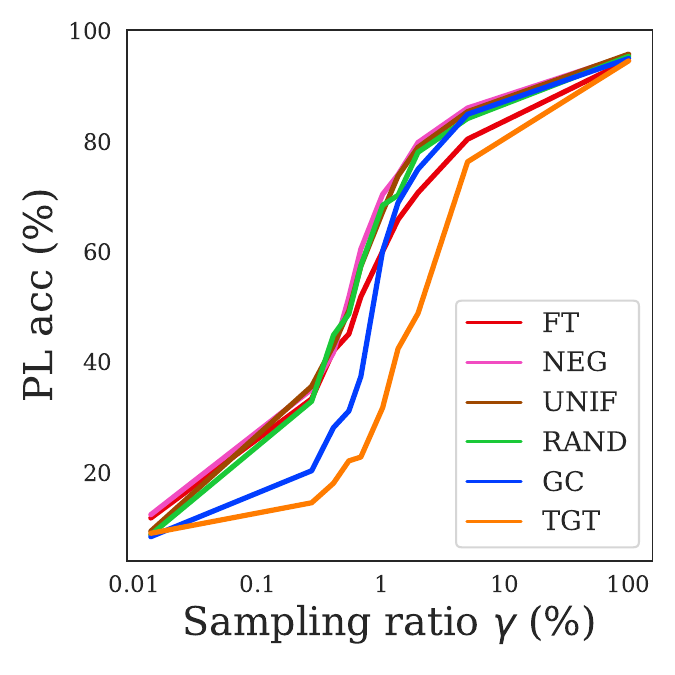}
  \caption{\footnotesize SVHN PL accuracy in \\CIFAR-100$\dtlarrow$STL-10-10\%.}
  \label{fig:pla_stl10_svhn}
     \end{subfigure}
     \begin{subfigure}{0.23\textwidth}
  \centering
  \includegraphics[width=0.95\linewidth]{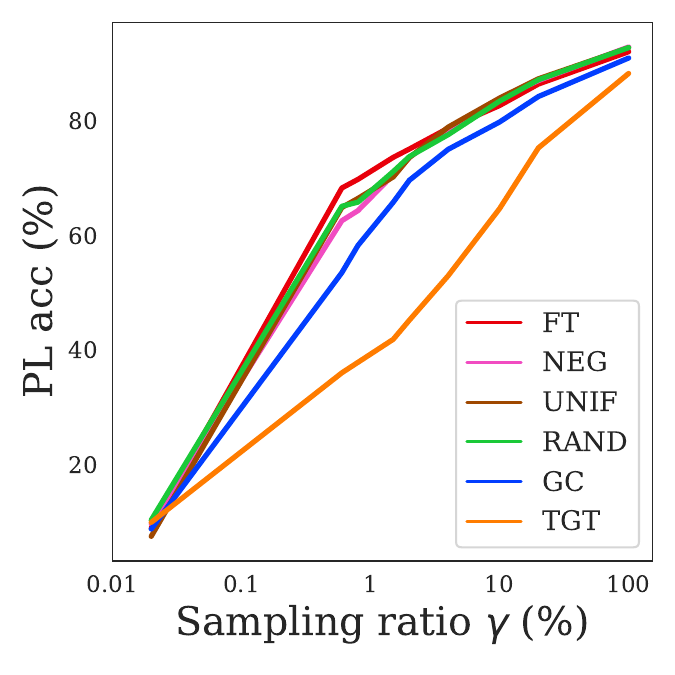}
  \caption{\footnotesize CIFAR-10 PL accuracy in \\CIFAR-100$\dtlarrow$STL-10-10\%.}
  \label{fig:pla_stl10_cif10}
     \end{subfigure}
        \caption{PL accuracy for different sampling ratio $\gamma$}
        \label{fig:pla}
\end{figure*}

Another possible unlearning might be conducted by distilling to a randomly initialized model. However, it fails to transfer target task performance because the teacher lacks the target task knowledge, as demonstrated $\tgtacc$ in \cref{tab:unlearning-distillation}.
This shows the target knowledge cannot be transferred by distillation. In a similar manner, other model compression methods such as pruning or quantization are not directly applicable for unlearning, as they are designed to best preserve the knowledge and cut down redundancy.

\subsection{Risks and limitations of GC unlearning}
A natural trade-off exists between knowledge disposal and the target task performance depending on the weight ($\lambda$) between knowledge retaining and unlearning loss. Notably, the proposed GC method demonstrated the most favorable trade-off (refer to \ref{fig:exp_unlearning_alpha}).
We further inspected the behavior of GC unlearning by setting an extreme case when the source and target are identical (see \cref{tab:curvature-maximization}). Increasing the weight on GC loss degrades the generalization performance while they equally achieve 100\% train accuracy. This is because the GC loss acts as a regularization that favors over-fitted solutions. Additionally, we found that increasing $\lambda$ results in higher loss curvature, which we measure by the trace of Hessian ($\textrm{tr}(\textrm{\bf H}_\theta)$).
\definecolor{Gray}{gray}{0.87} 
\newcolumntype{g}{>{\columncolor{Gray}}c}

\begin{table}[t]
\centering
\begin{tabular}{cc|gccc}
\toprule
Loss & $\lambda$             & {\small Train acc.} & {\small Test acc.} & {$\textrm{tr}(\textrm{\bf H}_\theta)$}  \\
\midrule
CE & 0.00 &100.00 & 76.13  & 66.2 \\
CE+GC & 0.10&100.00 & 72.18  & 226.2 \\
CE+GC & 0.15&100.00 & 71.12  & 314.4 \\
\bottomrule
\end{tabular}
\caption{GC loss training when source task and target task are identical (CIFAR-100). The norm of the curvature is measured by the trace of the Hessian matrix.}
\label{tab:curvature-maximization}

\end{table}

\begin{table}[t]
\centering
\resizebox{\linewidth}{!}{
\begin{tabular}{c|ccccc}
\toprule
\multirow{2}{2em}{Method} & \multirowcell{2}{$\Delta Acc_t$ \\vs TGT $\uparrow$} & \multirowcell{2}{$\Delta Acc_t$ \\vs TL $\uparrow$} & \multicolumn{3}{c}{$\Delta Acc_{pl}$ vs TL $\downarrow$} \\
\cmidrule(lr){4-6}
&&& {\footnotesize CIFAR100-10\%} & {\footnotesize STL10-10\%} & {\footnotesize SVHN-1\%} \\
\midrule
R18$\rightarrow$R18 & +3.98 & -31.83   & -39.31       & -10.32    & -34.71   \\
R50$\rightarrow$R18 & +3.88 & -31.93   & -39.06       &  -10.68   & -32.59   \\
\midrule
Reference & 35.12 & 70.93 & 68.15 & 53.28 & 61.97 \\
\bottomrule
\end{tabular}
}
\caption{DTL using knowledge distillation. R18 and R50 indicate ResNet-18/50.}
\label{tab:unlearning-distillation}
\end{table}

\subsection{The source accuracy cannot verify the knowledge disposal}
In \cref{sec:pla_effective}, we have briefly discussed why the source accuracy of the unlearned model cannot verify the success of knowledge disposal with \cref{tab:reject_srcacc}.

Moreover, there are additional reasons why the source accuracy cannot measure the degree of unlearning. Primarily, our work is motivated by the situation that the TL model is susceptible to be adapted to various tasks except for the target task. As a result, the source accuracy can be ignored for model evaluation in the first place.

Also, the source accuracy cannot represent the conformability of other tasks. We demonstrated the trade-off of the PL accuracy vs the source accuracy (\cref{fig:src_pla}) and the target accuracy vs the source accuracy (\cref{fig:tgt_src}). Those are plotted with the same data for \cref{fig:exp_unlearning_alpha}, in which we evaluated the performance with varying the unlearning loss and $\lambda$ in CIFAR-100$\dtlarrow$CIFAR-10-1\% experiment. As shown in \cref{fig:src_pla}, there the source accuracy has a low effect on the PL accuracy. Particularly, source accuracy is even negatively correlated with the PL accuracy because the Spearman correlation coefficient of them is $\rho_s=-0.505$.

As mentioned in \cref{tab:reject_srcacc}, the source accuracy is nearly uncorrelated to the target accuracy. It is found that except GC models, the source accuracy of every model is lower than 2\% with various ranges of target accuracy in \cref{fig:tgt_src} and the Spearman correlation coefficient measured on the target accuracy and source accuracy is $\rho=0.199$. 

\subsection{Complete results}

In \cref{fig:pla_cif10_stl10}, PL accuracy versus different sampling ratios on the STL-10 dataset is reported, whereas there are other analyses on different PL datasets after CIFAR-100$\dtlarrow$CIFAR-10-1\% in \cref{fig:nontarget_sampling}.

In addition, we also conducted the same analysis with \cref{fig:nontarget_sampling}  and \cref{fig:exp_unlearning_alpha} in the CIFAR-100$\dtlarrow$STL-10-10\% experiment.  Refer to \cref{fig:pla_stl10_cif100,fig:pla_stl10_svhn,fig:pla_stl10_cif10} and \cref{fig:alpha_tuning_stl}, respectively.

\subsection{Extended results on the membership inference attacks (MIAs)}

In  \cref{tab:MIA_summarized}, we have reported the success rate of various white-box MIAs -- Adv. Dist, Grad $w$, Grad $x$, and WB -- for our unlearning models. Furthermore, in \cref{tab:mia_full}, we provide extended results with more models and additional attack methods. We adapted the implementation of \cite{AdvDistance} and followed experimental details. Specifically, the two measures, \emph{AUROC} and \emph{accuracy}, are measured by varying the threshold for binary classification of membership inference. AUROC represents the trade-off of true positive ratio (TPR) and false positive ratio (FPR) and accuracy is the best accuracy among various thresholding. Also, the reported values are averaged success rates and corresponding standard deviations of 20 runs.

Here, we provide details on each attack method. First, we attacked the models with various black-box methods. The Softmax response (Softmax) method is the most naive attack method, which attacks the model with the softmax output from the assumption that the predicted output from training data will be more confident. Modified entropy (Mentr.) is a variant of the Softmax method because it measures the output's uncertainty and it determines the sample with low uncertainty to the training sample. The Loss (Loss) method finds out the training sample by measuring the loss function and decides the sample with a lower loss to the training data.

Also, we applied the white-box methods to our models. The gradient norm (Grad Norm) method calculates the $\ell_2$-norm of the gradient with respect to the model parameter and figures out the sample with a smaller norm to be a training sample. \cite{AdvDistance} proposes the adversarial distance (Adv. Dist) to measure the amount of perturbation of an example so that the model wrongly predicts the class of the sample in a white-box manner. 
 \cite{GradInvAttack} proposes a white-box attack method based on the gradient of the loss with respect to model weight (Grad $w$) and input (Grad $x$). The authors claim that the larger norm of both gradients indicates the sample is not a member of the training set.
 Though excluded in the main manuscript, the intermediate outputs (Int.Outs) method attacks a target model with the outputs of the final two layers as introduced in \cite{GradInvAttack}. White-box method (WB) \cite{WB} takes intermediate feature and gradient for MIA.

\end{document}